\documentclass[10pt,journal,compsoc]{IEEEtran}
\usepackage[numbers,sort&compress]{natbib}

\ifCLASSINFOpdf
\else
\fi
\usepackage{dblfloatfix}
\usepackage{amsmath}
\usepackage{graphicx}
\usepackage[colorlinks, citecolor=black,linkcolor=black,urlcolor=black]{hyperref}
\usepackage{subfig}
\usepackage{adjustbox}
\usepackage{float}
\usepackage{caption}
\usepackage{dblfloatfix} 

\usepackage[utf8]{inputenc} 
\usepackage[T1]{fontenc}    
\usepackage{hyperref}       
\usepackage{url}            
\usepackage{booktabs}       
\usepackage{amsfonts}       
\usepackage{nicefrac}       
\usepackage{microtype}      
\usepackage{lipsum}
\usepackage{natbib}
\usepackage{enumitem}
\usepackage{mathtools, nccmath}
\usepackage{algorithmic}
\usepackage{algorithm}
\usepackage{balance}
\usepackage{bm}
\usepackage{xcolor}
\usepackage{amssymb}
\usepackage{pifont}

\usepackage{fancyhdr}
\usepackage{kantlipsum}
\usepackage{eso-pic}

\begin{document}


%
\title{Semi-Supervised Clustering with Inaccurate Pairwise Annotations}

\author{Daniel~Gribel,
        Michel~Gendreau,
        and~Thibaut~Vidal
\IEEEcompsocitemizethanks{\IEEEcompsocthanksitem D. Gribel and T. Vidal are with the Department
of Informatics, Pontif\'icia Universidade Cat\'olica do Rio de Janeiro, Rio de Janeiro, Brazil, 22451-900.\protect\\
E-mail: dgribel@inf.puc-rio.br, thibaut.vidal@cirrelt.ca
\IEEEcompsocthanksitem M. Gendreau and T. Vidal are with CIRRELT and the Department of Mathematics and Industrial Engineering, Polytechnique Montr\'eal, P.O. Box 6079. Succ. Centre-ville, Montr\'eal, QC, H3C 3A7 Canada.\protect\\
Email: michel.gendreau@cirrelt.net
}
}

\IEEEtitleabstractindextext{%
\begin{abstract}
Pairwise relational information is a useful way of providing partial supervision in domains where class labels are difficult to acquire. This work presents a clustering model that incorporates pairwise annotations in the form of \textit{must-link} and \textit{cannot-link} relations and considers possible annotation inaccuracies (i.e., a common setting when \textit{experts} provide pairwise supervision). We propose a generative model that assumes Gaussian-distributed data samples along with \textit{must-link} and \textit{cannot-link} relations generated by stochastic block models. We adopt a maximum-likelihood approach and demonstrate that, even when supervision is weak and inaccurate, accounting for relational information significantly improves clustering performance. Relational information also helps to detect meaningful groups in real-world datasets that do not fit the original data-distribution assumptions. Additionally, we extend the model to integrate prior knowledge of experts' accuracy and discuss circumstances in which the use of this knowledge is beneficial.
\end{abstract}

}

\maketitle

\IEEEdisplaynontitleabstractindextext

%
\IEEEpeerreviewmaketitle

\IEEEraisesectionheading{\section{Introduction}\label{sec:Introduction}}
\IEEEPARstart{D}{ata} clustering aims at systematically grouping a set of  data samples such that samples with similar features are placed within the same cluster, whereas samples with a certain degree of separability are allocated to different clusters. Although clustering is an unsupervised-learning task, situations exist in which partial annotations are given with the dataset \citep{Schwenker2014}, leading to semi-supervised models.

\begin{figure}[!b]
  \vspace*{-0.7cm}
  \centering
    \captionsetup{justification=centering}
    \subfloat[Unsupervised clustering]{{\includegraphics[width=4.4cm]{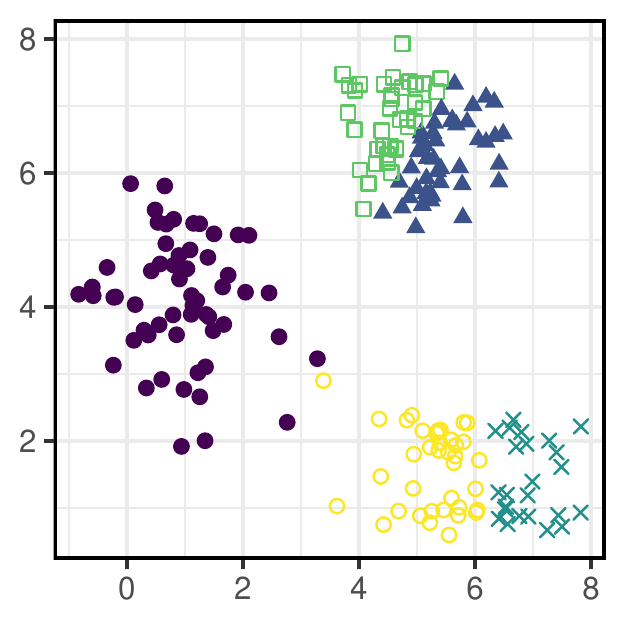}}}%
    \hspace{0.0em}
    \subfloat[Pairwise-clustering]{{\includegraphics[width=4.4cm]{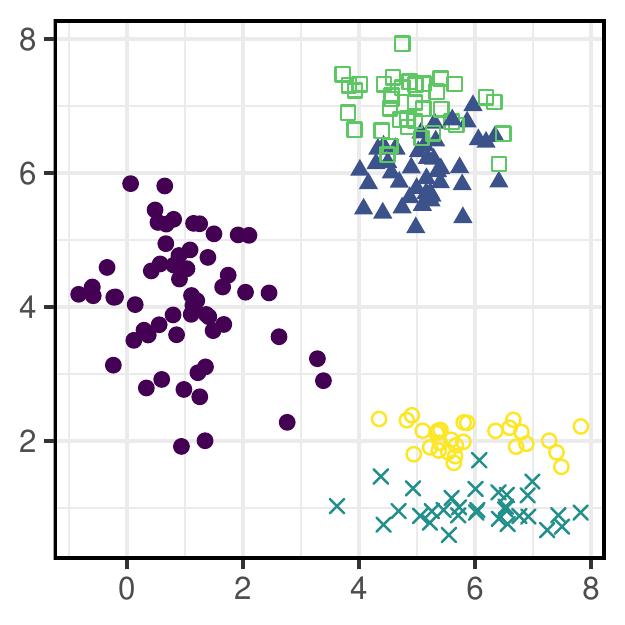}}}%
    \caption{Different partitions in a mixture of Gaussians.}
    \label{fig:Example1}
\end{figure}

In particular, relational information in the form of pairwise constraints sees regularly use: \textit{must-link} constraints state that a pair of data samples should belong to the same cluster, whereas \textit{cannot-link} constraints separate pairs of data samples into different groups.
Relational information is usually inferred through similarity measures or provided by domain experts and can provide semi-supervision in domains where it is difficult, time-consuming, or expensive to accurately measure the actual classes \cite{Basu2004,Basu2008}.
Incorporating relational supervision can bring significant benefits. Figs.~\ref{fig:Example1}(a)\ and \ref{fig:Example1}(b), for example, compare clustering solutions obtained without and with semi-supervised learning, respectively,
on a dataset with 200 samples and 600 random pairwise annotations. In this example, the relational information  guides the clustering algorithm out of a local minimum of the unsupervised model toward a solution close to the ground-truth.

The present study focuses on the use of relational information in clustering. We consider a regime in which \textit{experts} or automated procedures provide pairwise annotations indicating whether pairs of observations belong to the same group or not.
This regime presents two notable characteristics: First, the annotators are not entirely accurate, so the relational information is given with some level of trust. Second, they have a limited work capacity, so only a small amount of pairwise relational information is available.

Some previous works  focused on semi-supervised clustering settings with relational information, and especially on
variants of the minimum sum-of-squares clustering (MSSC) model with additional pairwise constraints \citep{wagstaff2001,Basu2004,bilenko2004,Pelleg2007}. The K-means algorithm \citep{Hartigan1979} is a well-known local optimizer of this formulation, and successive improvements of this solution method have been proposed over the years \citep{Likas2003,Vassilvitskii2006,Ordin2015,Gribel2019}.
However, most of these studies incorporate pairwise information in classical search algorithms such as K-means through additional ad hoc constraints or soft penalty factors. By doing so, these approaches lack a probabilistic interpretation and may fail in the presence of noisy and scarce supervision due to erroneous binding constraints. Similarly, soft penalties depend largely  on good parameter choices.

To cope with these issues, we introduce a maximum-likelihood approach for a generative model that assumes that data samples are generated by spherical Gaussian distributions. The \textit{must-link} and \textit{cannot-link} constraints occur between a pair of data samples with probabilities that depend only on the groups that contain the samples. To model the presence of \textit{must-link} and \textit{cannot-link} relations, we assume graphs generated by stochastic block models (SBMs) and integrate prior beliefs to represent possible knowledge of the experts' accuracy. We further propose efficient solution techniques for this model based on the HG-means approach \citep{Gribel2019}, a state-of-the-art algorithm for the MSSC model that enhances the classical K-means approach through successive restarts from promising starting points obtained by recombination.

Finally, we conduct extensive computational experiments by applying the proposed model to synthetic and real-world datasets to measure how relational information affects clustering. We show that pairwise annotations can significantly improve clustering performance, even when given only a small amount of imperfect supervision. Incorporating pairwise annotations can also reveal clustering structures not detected by unsupervised approaches, as demonstrated on a real-world dataset. Finally, we show that incorporating prior knowledge regarding the experts' accuracy further guides the clustering process toward more accurate partitioning.

\section{Related Works}
\label{sec:RelatedWorks}
Several clustering formulations have been proposed to exploit pairwise information, e.g., based on expectation-maximization (EM) \citep{Shental2004,Basu2008}, spectral clustering \citep{Li2009,Wang2014}, or affinity propagation \citep{Givoni2009,Arzeno2015}.

Some previous works have adapted the MSSC objective function to incorporate pairwise constraints. \citet{wagstaff2001} proposed a variant of the K-means algorithm that imposes that no constraint is violated. However, such a model may fail to find a feasible solution. \citet{Basu2004} and \citet{Hiep2016} included a penalty term that is either uniform or proportional to the distance between samples in the dataset. \citet{bilenko2004} studied the MSSC with pairwise constraints and proposed a metric-adaptive penalty factor according to which the penalty of a violated \textit{must-link} is greater for two distant samples than for two close samples. An analogous notion holds for \textit{cannot-links}. \citet{Pelleg2007} also explored an extension of K-means in which the violated pairwise constraints are tentatively solved by moving one cluster's center toward another one to change the regions of the feature space covered by the clusters and thereby satisfy the constraints.

\citet{Bai2020} included supervision from different sources (pairwise constraints, positive labeling, and negative labeling) in a pairwise relational matrix representation. For the resulting optimization problem, the authors proposed eigenvalue decomposition methods that jointly maximize within-cluster similarity and the consensus among the different supervision. \citet{Shental2004} modified the Gaussian mixture model (GMM) likelihood to incorporate \textit{must-link} and \textit{cannot-link} constraints and designed an EM algorithm with tailored update rules to handle these constraints. \textit{Must-link} constraints are handled by collapsing data samples through transitive closure, whereas \textit{cannot-links} are described through Markov networks. However, erroneous pairwise relations can strongly affect the results of the algorithm.

All these approaches  are adaptations of the MSSC formulation and the GMM to cluster data samples with additional pairwise constraints. Thus, the relational information is incorporated into the formulation to find a partition (e.g., by using the violation of pairwise constraints as penalty factors). An alternative way to jointly consider the data features and the relational information that we adopt in this paper consists in modeling the observed data from a probabilistic perspective. According to this perspective, the features and pairwise relations are assumed to come from a generative model, which is fit to the data.

SBMs \citep{Holland1983, Nowicki2001} are general classes of random graph models commonly used to detect clusters based only on relational information. When such graphs have some structure,
fitting the parameters of a SBM to empirical graphs is widely adopted to reveal blocks (clusters). In the canonical form of SBMs, the expected number of edges between two samples is  determined solely by the blocks to which they belong. In this way, samples within each block are statistically equivalent in terms of their connectivity patterns. SBMs are regularly used to recover meaningful information from complex graphs and are also a natural modeling choice for community detection. The surveys of \citet{Abbe2017} and of \citet{Lee2019} discuss key concepts and solution algorithms in stochastic block modeling. Different types of algorithms can be used to fit SBMs based on Markov chain Monte Carlo  approaches \citep{Mcdaid2013,Nowicki2001,Peixoto2019}, variational inference \citep{Wang2017,Airoldi2008}, belief propagation \citep{Decelle2011}, spectral clustering \citep{Lei2015,Qin2013,Rohe2011}, or semidefinite programming \citep{Cai2015,Chen2012}, among others.

Previous studies have proposed to extend SBMs to consider additional data features (also referred to as meta-data). 
\citet{Stanley2019} presented a probabilistic model that combines relational information and data features within a ``soft membership'' formulation. In the derived model, SBM probabilities define the graph connectivity, and Gaussian parameters describe the features. The authors employ EM algorithms to maximize the resulting likelihood function. Although  EM works well for estimating the Gaussian parameters, the computation of the conditional distributions to get the assignment probabilities is not tractable with SBMs~\cite{Daudin2008}. They therefore use a variational approach that optimizes a lower bound of the SBM likelihood function. Several experiments on link prediction and collaborative filtering on biological datasets have been  reported.

\citet{Contisciani2020} introduced a probabilistic model for community detection in multi-layer graphs, combining sample features with relational information, where the sample features are categorical. Each category has a probability of being observed in a community, while a SBM variant serves to model relational information. Thus, the proposed model includes two independent likelihood functions and assumes conditional independence of the observed features and networks. Given that each likelihood may  differ in magnitude, the authors propose using a weight---tuned by cross-validation---that inclines the model toward one of the formulations. As a consequence, this approach diverges from a maximum-likelihood perspective.

The techniques described above represent fundamental advances in semi-supervised models and methods. However, they either involve firm constraints and cannot handle imprecise annotations, or they depend on soft penalty factors that are hard to calibrate. Finally, they do not directly derive from a maximum-likelihood interpretation. In what follows, we fill this gap and propose principled probabilistic models to set experts' annotations.

\section{Proposed Model}
\label{sec:Model}
In the pairwise-constrained clustering problem, we are given a set $\bm{X} = \{\bm{x}_1, \dots, \bm{x}_N\}$ with $N$ data samples in~$\mathbb{R}^D$ along with a symmetric adjacency matrix $\bm{A} \in \mathbb{N}^{N, N}$ representing some relational information between the data samples, where the entry $A_{ij}$ indicates the number of existing edges between data samples $\bm{x}_i$ and $\bm{x}_j$. Typically, pairwise constraints express some hard association. For example, they indicate whether two samples should be assigned to the same cluster or to different clusters. We then aim to
 partition the data samples into $K$ disjoint clusters $\mathcal{C} = \{C_1, \dots, C_K\}$ with the goal of optimizing a given clustering criterion. One way to formalize this problem is to define a likelihood function and fit this function's parameters to the observed data.\\

\noindent \textbf{Gaussian Mixture Model}. The Gaussian Mixture Model (GMM) is a widely used probabilistic model that assumes data samples generated by a finite number of Gaussian distributions. The model parameters are the mean points and the covariance matrices of each cluster, and the assignment of samples to clusters is a latent variable. In this work, we explore the hard-membership version of the GMM, which assumes that each data sample is assigned  to exactly one cluster, so that the latent assignment variable becomes binary. It is well known that maximizing the likelihood of the hard-membership GMM (also referred to as the MSSC) approximates the ordinary GMM, and algorithms such as  K-means act as a variational expectation-maximization in the GMM
\cite{Lucke2019}. The log-likelihood function for the hard-membership GMM can be calculated as per \citet{Bishop2006}:
\begin{equation}
\log P(\bm{X} | \bm{\mu}, \bm{\Sigma}, \bm{Z}) = \sum_{i}^{N} \sum_{r}^{K} z_{ir} \log \mathcal{N} (\bm{x}_i | \bm{\mu}_r, \bm{\Sigma}_r),
\end{equation}
where $\bm{Z} \in \mathbb{R}^{N, K}$ is the binary cluster indicator such that each entry $z_{ir} \in \{0,1\}$ takes the value $1$ if and only if sample~$i$ belongs to cluster $r$, so $\sum_{r}^{K}z_{ir} = 1\ \forall\ i \in \{1, \dots, N\}$.
The variables $\bm{\mu} = \{\bm{\mu}_1, \dots, \bm{\mu}_K \}$ and $\bm{\Sigma} = \{\bm{\Sigma}_1, \dots, \bm{\Sigma}_K \}$ contain the means and covariances of the Gaussian components, respectively. For the special case of spherical GMMs, $\bm{\Sigma}_r = \sigma_r^2 \mathbf{I}\ \forall\ r \in \{1, \dots, K\}$, and the log-likelihood function may be expressed as

\begin{equation}
\label{EqGmm01}
\begin{split}
        & \log P(\bm{X} | \bm{\mu}, \bm{\sigma}, \bm{Z}) = \sum_{i}^{N} \! \sum_{r}^{K} z_{ir} \log \left ( \frac{e^{ -\left \| \bm{x}_i - \bm{\mu}_r \right \|^2/{2 \sigma_r^2}}}{(2 \pi)^{D/2} \sigma_r^D} \right ) \\
        & \hspace*{-0.06cm}= \hspace*{-0.06cm}\sum_{i}^{N} \! \sum_{r}^{K} z_{ir}\hspace*{-0.08cm} \left (\hspace*{-0.08cm} - \frac{\left \| \bm{x}_i - \bm{\mu}_r \right \|^2}{2 \sigma_r^2} - \frac{D}{2} \log(2 \pi) - D \log(\sigma_r)\hspace*{-0.08cm} \right ) \hspace*{-0.1cm}.
\end{split}
\end{equation}

When the assignments are fixed, as seen in \citet{Bishop2006}, the maximum of this log-likelihood function occurs when

\begin{equation}\label{MuDerivation}
\bm{\hat{\mu}}_r = \frac{ \sum_{i}^{N} z_{ir} \bm{x}_i }{ \sum_{i}^{N} z_{ir} }
\end{equation}

\noindent and

\begin{equation}\label{SigmaDerivation}
\hat{\sigma}_r^2 = \frac{\sum_{i}^{N} \sum_{r}^{K} z_{ir} \! \left \| \bm{x}_i - \bm{\mu}_r \right \|^2}{2D \sum_{i}^{N} z_{ir}}.
\end{equation}

Therefore, suppressing the constant $D \log(2 \pi)$ and rearranging terms, we obtain:

\begin{equation}
\label{EqGmm02}
\hspace*{-0.1cm}\log P(\bm{X} | \bm{Z}) \hspace*{-0.08cm} \propto \hspace*{-0.08cm} - \frac{1}{2} \sum_{i}^{N} \! \sum_{r}^{K} z_{ir} \hspace*{-0.08cm} \left ( \hspace*{-0.08cm} \frac{\left \| \bm{x}_i \hspace*{-0.08cm} - \hspace*{-0.08cm} \bm{\hat{\mu}}_r \right \|^2}{\hat{\sigma}_r^2} - 2D \hspace*{-0.05cm} \log(\hat{\sigma}_r) \hspace*{-0.1cm} \right ) \hspace*{-0.1cm}.
\end{equation}

\noindent \textbf{Stochastic Block Models.} The likelihood function of a GMM is a well-known clustering formulation when data samples have continuous features. To incorporate pairwise constraints into our semi-supervised setting, we now briefly review SBMs, a family of probabilistic models used to detect structure in graphs, and then proceed toward a unified formulation that considers both feature-based samples and relational information.

In its most fundamental form, a SBM considers $N$ data samples and $K$ groups, where each sample is originally assigned to one group. Then, we assume undirected edges placed between two samples at random with expected value~$\omega_{rs}$ that depends only on groups $r$ and $s$ to which the data samples belong \cite{Newman2016}. Finding the latent membership of data samples in a SBM corresponds to finding the block-model parameters $\bm{\Omega}$ that best fit an observed graph \cite{Abbe2017}.
For an observed adjacency matrix~$\bm{A} \in \mathbb{N}^{N, N}$ representing a graph with $m$ possibly weighted edges, the log-likelihood function of the SBM can be expressed as per \citet{Karrer2011}:
\begin{equation}\label{SBM}
\log P(\bm{A} | \bm{\Omega}, \bm{Z}) = \frac{1}{2} \sum_{rs}^{K}\hspace*{-0.03cm} \sum_{ij}^{N} \left ( A_{ij} \log(\omega_{rs}) - \omega_{rs}  \right ) z_{ir} z_{js},
\end{equation}
where parameters $\bm{Z}$ and $\bm{\Omega}$ are the latent variables, with $\bm{Z} \in \mathbb{R}^{N, K}$ being the binary cluster indicator, and $\omega_{rs}$ being an entry of $\bm{\Omega}$ that represents the expected number of edges between any two samples in clusters~$r$ and~$s$. If we fix the assignment $\bm{Z}$ in Equation~\eqref{SBM}, then the maximum-likelihood values of $\omega_{rs}$ can be found by differentiation:

\begin{equation}\label{OmegaDerivation}
\hat{\omega}_{rs} = \frac{ m_{rs}}{ n_{r} n_{s} },
\end{equation}

\noindent where \smash{$m_{rs} = \sum_{ij}^{N} A_{ij} z_{ir} z_{js}$}
is the number of edges between clusters $r$ and $s$, and \smash{$n_{r} = \sum_{i}^{N} z_{ir}$} is the number of samples in cluster $r$. Using the closed form of $\bm{\Omega}$ from  Equation~\eqref{OmegaDerivation}, the log-likelihood of the SBM can be rewritten as
\begin{equation}\label{DCSBMDerivation}
\begin{split}
\log P(\bm{A} | \bm{Z}) & = \frac{1}{2} \sum_{rs}^{K} \sum_{ij}^{N} \left ( A_{ij} \log(\hat{\omega}_{rs}) - \hat{\omega}_{rs}  \right ) z_{ir} z_{js}\\
& = \frac{1}{2} \sum_{rs}^{K} \sum_{ij}^{N}\left ( A_{ij} \log \left ( \frac{m_{rs}}{n_r n_s} \right )  \right ) z_{ir} z_{js}.
\end{split}
\end{equation}

\subsection{Experts' Annotations Setting}\label{ExpertsAnnotation}
Our proposed generative model considers a set $\bm{X} = \{\bm{x}_1, \dots, \bm{x}_N\}$ of $N$ samples in $\mathbb{R}^D$, along with two independent graphs $\bm{A}^{+}$ and $\bm{A}^{-}$ that represent the \textit{must-link} and \textit{cannot-link} relations in the form of adjacency matrices. These annotations are produced by experts on a subset of sample pairs. The complete generative process can be described as follows:

\begin{itemize}
    
    \item For each $i \in \{1, \dots, N\} $:        
        \begin{itemize}            
            \item Pick a Gaussian component $r \in \{1, \dots, K\}$ with uniform probability $1/K$, and set $\hat{y}_{i} = r$ as the ground-truth;        
            \item Generate a $D$-dimensional sample $\bm{x}_i$ from component~$r$:    
                        \begin{equation}\label{EqGaussExpert}
                         \bm{x}_i \sim \mathcal{N}(\bm{\mu}_r, \sigma_r^2).
                        \end{equation}
        \end{itemize}
     
    \item For each sample pair $(\bm{x}_i, \bm{x}_j)$ selected independently and with uniform probability, an expert labels the pair as a \textit{must-link} or a \textit{cannot-link} relation
    according to a Bernoulli distribution, which is defined based on the groups to which the samples belong:
                
        \begin{equation}\label{EqPoissonExpert}
        \begin{cases} 
        A_{ij}^{+} = \text{Bernoulli}(p_{\hat{y}_{i} \hat{y}_{j}}), \\
        A_{ij}^{-} = 1 - A_{ij}^{+},
        \end{cases}
        \end{equation}
                        
        \noindent in which $p_{rs} \in [0, 1]$ is the probability of marking a pair of samples as a \textit{must-link} given that the samples belong to groups~$r$ and~$s$. Analogously, $1 - p_{rs}$ is the probability of marking a pair of samples in groups $r$ and~$s$ as a \textit{cannot-link}. Typically, 
        $p_{rr} \geq p_{rs}$ when $r \neq s$.
\end{itemize}

Assuming that $m$ annotations are generated independently and uniformly between sample pairs, the expected number of \textit{must-link} edges between an arbitrary pair of samples from groups $r$ and $s$ can be estimated as $\beta p_{rs}$, with $\beta = 2m/[n(n+1)]$. Similarly, the expected number of \textit{cannot-link} edges is $\beta (1-p_{rs})$. We can thus model the experts' annotations setting by using two stochastic block models with matrices $\bm{\Omega}^{+}$ and $\bm{\Omega}^{-}$ for the \textit{must-link} and \textit{cannot-link} graphs, respectively. In this case, $\omega_{rs}^{+} \sim \beta p_{rs}$ and $\omega_{rs}^{-} \sim \beta (1-p_{rs})$. Since multiple experts provide annotations with replacement, we obtain Poisson-distributed matrices $\bm{\Omega}^{+}$ and $\bm{\Omega}^{-}$. Note, however, that the 
two  graphs produced are not independent because a ``failure'' in a Bernoulli trial generates an edge in the \textit{cannot-link} graph. Nonetheless, independence holds between annotations because of sample pair selections with replacement, such that we can reasonably approximate the experts' annotations by two independent SBMs with parameters $\bm{\Omega}^{+}$ and $\bm{\Omega}^{-}$ for \textit{must-link} and \textit{cannot-link} relations, respectively:

\begin{equation}
\begin{split}
\label{Likelihood}
P(\bm{X}, \bm{A}^{+}, \bm{A}^{-} | \bm{\mu}, \bm{\sigma}, \bm{\Omega}, \bm{Z} ) = &\, P(\bm{X} | \bm{\mu}, \bm{\sigma}, \bm{Z} )  \\
&\times P(\bm{A}^{+} | \bm{\Omega}^{+}, \bm{Z} )\\
& \times P(\bm{A}^{-} | \bm{\Omega}^{-}, \bm{Z} ).
\end{split}
\end{equation}

Hereinafter, we consider $\mathcal{L}(\cdot) = \log P(\cdot)$ to refer to a log-likelihood function, $\bm{A} = \{\bm{A}^{+}, \bm{A}^{-}\}$ to represent the \textit{must-link} and \textit{cannot-link} graphs, and $\bm{\Omega} = \{\bm{\Omega}^{+}, \bm{\Omega}^{-}\}$ to represent the two SBM matrices. Thus, the resulting log-likelihood function is

\begin{equation}
\begin{split}
\label{LogLikelihood}
\mathcal{L} (\bm{X}, \bm{A} | \bm{\mu}, \bm{\sigma}, \bm{\Omega}, \bm{Z}) \! \propto
& \! - \! \sum_{i}^{N} \! \sum_{r}^{K} \! \left ( \! \frac{\left \| \bm{x}_i \! - \! \bm{\mu}_r \right \|^2}{\sigma_r^2} \! + \! 2D \! \log(\sigma_r) \! \right ) \! z_{ir}\\
& + \! \sum_{rs}^{K} \! \sum_{ij}^{N} \! \left ( A_{ij}^{+} \log(\omega_{rs}^{+}) - \omega_{rs}^{+} \right ) z_{ir} z_{js} \\
& + \! \sum_{rs}^{K} \! \sum_{ij}^{N} \! \left ( A_{ij}^{-} \log(\omega_{rs}^{-}) - \omega_{rs}^{-} \right ) z_{ir} z_{js},
\end{split}
\end{equation}

\noindent where we removed the constant $\frac{1}{2}$ in front of all terms. The variables $\bm{\mu}_r$ and $\sigma_r$ are obtained from Equations~\eqref{MuDerivation} and~\eqref{SigmaDerivation}, respectively, and $\omega_{rs}^{+}$ and $\omega_{rs}^{-}$ are obtained from Equation~\eqref{OmegaDerivation}. As a consequence, we can write this log-likelihood as
\begin{equation}
\mathcal{L} (\bm{X}, \bm{A} | \bm{Z})  = \log P(\bm{X} | \bm{Z}) + \log P(\bm{A}^{+} | \bm{Z}) + \log P(\bm{A}^{-} | \bm{Z}).
\end{equation}

\subsection{\textit{Prior} Knowledge of Experts' Accuracy}
Although the SBMs are used to infer partitions of any structure, it is common in practice to have an estimate of the experts' accuracy. In some circumstances, we may reasonably assume to have pre-evaluated the experts' accuracy before the annotation procedure. Consequently, we propose an extension of model~\eqref{LogLikelihood} that incorporates a \textit{prior} belief regarding the accuracy of annotations. We first consider the maximum posterior estimate of parameters $\bm{\Omega}$ and $\bm{Z}$ in the SBM:

\begin{equation}
P(\bm{\Omega}, \bm{Z} | \bm{A}) \propto P(\bm{A} | \bm{\Omega}, \bm{Z}) P(\bm{\Omega}, \bm{Z}),
\end{equation}

\noindent where the joint prior distribution is

\begin{equation}
P(\bm{\Omega}, \bm{Z}) = P(\bm{\Omega} | \bm{Z}) P(\bm{Z}),
\end{equation}

\noindent and we assume that $P(\bm{Z})$ has the same probability for any assignment $\bm{Z}$ and thus is treated as a constant.
As in \citet{Peixoto2019}, we opt for the following form of a prior function:
\begin{equation}
\begin{split}
P(\bm{\Omega} | \bm{Z}) & = \prod_{r \leq s} \lambda_{rs}(\bm{Z}, p) \enspace e^{- \lambda_{rs}(\bm{Z}, p) \omega_{rs}}\\
& = \prod_{rs} \left ( \lambda_{rs}(\bm{Z}, p) \enspace e^{- \lambda_{rs}(\bm{Z}, p) \omega_{rs}} \right )^{\frac{1}{2}(1 + \delta_{rs})},
\end{split}
\end{equation}
where $\delta_{rs}$ is the Kronecker delta and $1/\lambda_{rs}(\bm{Z}, p)$ is the expected (mean) value in the exponential distribution. Although the experts' accuracy is fixed (for example, $p = 90\%$), the values we choose for our priors depends on~$\bm{Z}$, since the assignment choices impact the size of the clusters and therefore the expected total number of annotations. This dependence occurs because SBMs have two sets of parameters, making our prior distribution conditioned on~$\bm{Z}$. $\lambda_{rs}(\bm{Z}, p)$ can be expressed as:
\begin{equation}
\lambda_{rs}(\bm{Z}, p) =
\begin{cases} 
1/f_{\text{IN}} (\bm{Z}, p) & \text{if $r = s$}\\
1/f_{\text{OUT}} (\bm{Z}, p) & \text{otherwise},
\end{cases}
\end{equation}
where, for a given $\bm{Z}$ and~$p$, $f_{\text{IN}} (\bm{Z}, p)$ represents a prior knowledge of the expected number of edges between two samples in the same group. Similarly, $f_{\text{OUT}} (\bm{Z}, p)$ represents the prior expected number of edges between two samples in different groups. For the sake of brevity, we will use the short form $\lambda_{rs} = \lambda_{rs}(\bm{Z}, p)$ in the remainder of this section. Since we have two graphs, we use $\lambda_{rs}^{+}$ and $\lambda_{rs}^{-}$ to refer to our priors in the \textit{must-link} and \textit{cannot-link} graphs, respectively, along with functions $f^{+}_{\text{IN}} (\bm{Z}, p)$, $f^{+}_{\text{OUT}} (\bm{Z}, p)$, $f^{-}_{\text{IN}} (\bm{Z}, p)$, and $f^{-}_{\text{OUT}} (\bm{Z}, p)$. Suitable values for these functions are discussed later in this section. This leads to the following posterior distribution:
\begin{equation}
\begin{split}
\label{Posterior}
\hspace*{-0.1cm} P(\bm{\Omega}, \bm{Z} | \bm{A}) & \propto P(\bm{A}^{+} | \bm{\Omega}^{+}, \bm{Z} ) \prod_{rs} \hspace*{-0.08cm} \left ( \lambda_{rs}^{+} e^{- \lambda_{rs}^{+} \omega_{rs}^{+}} \right )^{\hspace*{-0.06cm}\frac{1}{2}(1 + \delta_{rs})}
\\
& \times P(\bm{A}^{-} | \bm{\Omega}^{-}, \bm{Z} ) \prod_{rs} \hspace*{-0.08cm} \left ( \lambda_{rs}^{-} e^{- \lambda_{rs}^{-} \omega_{rs}^{-}} \right )^{\hspace*{-0.06cm}\frac{1}{2}(1 + \delta_{rs})}\hspace*{-0.08cm},\hspace*{-0.08cm}
\end{split}
\end{equation}
and therefore to the following log-posterior with the observed features~$\bm{X}$:
\begin{equation}
\begin{split}
\label{LogPosterior}
\mathcal{L} (\bm{\mu}, \bm{\sigma}, \bm{\Omega}, \bm{Z} | \bm{X}, \bm{A}) \! \propto
& \! - \! \sum_{i}^{N} \! \sum_{r}^{K} \! \left ( \! \frac{\left \| \bm{x}_i \! - \! \bm{\mu}_r \right \|^2}{\sigma_r^2} \! + \! 2D \! \log(\sigma_r) \! \right ) \! z_{ir}\\
& + \! \sum_{rs}^{} \! \sum_{ij}^{} \left ( A_{ij}^{+} \log(\omega_{rs}^{+}) - \omega_{rs}^{+} \right ) z_{ir} z_{js} \\
& + \! \sum_{rs}^{} \! \sum_{ij}^{} \left ( A_{ij}^{-} \log(\omega_{rs}^{-}) - \omega_{rs}^{-} \right ) z_{ir} z_{js} \\
& + \! \sum_{r}^{} \log(\lambda_{rr}^{+} \lambda_{rr}^{-}) - \lambda_{rr}^{+} \omega_{rr}^{+} - \lambda_{rr}^{-} \omega_{rr}^{-} \\
& + \! \sum_{rs}^{} \log(\lambda_{rs}^{+} \lambda_{rs}^{-}) - \lambda_{rs}^{+} \omega_{rs}^{+} - \lambda_{rs}^{-} \omega_{rs}^{-}.
\end{split}
\end{equation}

\noindent In Equation~(\ref{LogPosterior}), we used a constant prior for the mixture of Gaussians, and therefore only take the likelihood into account. The last two summations come from the exponential priors. 
The optimal value of $\omega_{rs}$ in the posterior log-likelihood can then be estimated by differentiation:

\begin{equation}
\hat{\omega}_{rs} =
\begin{cases} 
m_{rs}/({n_r n_s + 2 \lambda_{rs}}) & \text{if $r = s$}\\
m_{rs}/({n_r n_s + \lambda_{rs}}) & \text{otherwise},
\end{cases}
\end{equation}

\noindent where we substitute $\hat{\omega}_{rs}$ with the corresponding parameter $\omega_{rs}^{+}$ or $\omega_{rs}^{-}$, depending on the graph (likewise for $m_{rs}$ and~$\lambda_{rs}$).\\

\noindent \textbf{Parametrization of the priors.}
%
In the \textit{must-link} graph, we can use $\lambda_{rs}^{+} = 1/f^{+}_{\text{IN}}(\bm{Z}, p)$ for $r = s$, and $\lambda_{rs}^{+} = 1/f^{+}_{\text{OUT}}(\bm{Z}, p)$ for $r \neq s$. In the \textit{cannot-link} graph, $\lambda_{rs}^{-} = 1/f^{-}_{\text{IN}}(\bm{Z}, p)$ for $r = s$, and $\lambda_{rs}^{-} = 1/f^{-}_{\text{OUT}}(\bm{Z}, p)$ otherwise. Due to the experts' annotations setting, the following relationship holds between $f^{+}_{\text{IN}}(\bm{Z}, p)$ and $f^{+}_{\text{OUT}}(\bm{Z}, p)$:
\begin{equation}\label{eq:RelationMust}
f^{+}_{\text{IN}}(\bm{Z}, p) = \frac{p}{1-p} f^{+}_{\text{OUT}}(\bm{Z}, p).
\end{equation}
Analogously, for the \textit{cannot-link} graph, we have
\begin{equation}\label{eq:RelationCannot}
f^{-}_{\text{IN}}(\bm{Z}, p) = \frac{1-p}{p} f^{-}_{\text{OUT}}(\bm{Z}, p).
\end{equation}

The number of pairs within and between groups given by~$\bm{Z}$, and the number of \textit{must-link} annotations $m^{+}$ and \textit{cannot-link} annotations $m^{-}$ also lead to the following relations:
\begin{equation} \label{eq:EstimatePriorMust}
f^{+}_{\text{IN}}(\bm{Z}, p) \sum_{r}^{} \frac{n_r (n_r\!+\!1)}{2} + f^{+}_{\text{OUT}}(\bm{Z}, p) \sum_{r < s}^{} n_r n_s = m^{+},
\end{equation}
\begin{equation} \label{eq:EstimatePriorCannot}
f^{-}_{\text{IN}}(\bm{Z}, p) \sum_{r}^{} \frac{n_r (n_r\!+\!1)}{2} + f^{-}_{\text{OUT}}(\bm{Z}, p) \sum_{r < s}^{} n_r n_s = m^{-},
\end{equation}
where $n_r = \sum_{i}^{} z_{ir}$ is the number of samples in group~$r$. Then, combining Equations~(\ref{eq:RelationMust}--\ref{eq:EstimatePriorCannot}) leads to:
\begin{equation}
\small\label{eq:PriorMust}
f^{+}_{\text{IN}}(\bm{Z}, p) = m^{+}  \hspace*{-0.1cm}  \left/ \hspace*{-0.02cm} \left( \sum_{r}^{} \frac{n_r (n_r\!+\!1)}{2} + \frac{(1\!-\!p)}{p} \sum_{r < s}^{} n_r n_s \right )\right.\hspace*{-0.08cm},
\end{equation}
\begin{equation}
\small\label{eq:PriorCannot}
f^{-}_{\text{IN}}(\bm{Z}, p) = m^{-} \hspace*{-0.1cm} \left/ \hspace*{-0.02cm} \left ( \sum_{r}^{} \frac{n_r (n_r + 1)}{2} + \frac{p}{(1-p)} \sum_{r < s}^{} n_r n_s \right )\right.\hspace*{-0.08cm}.
\end{equation}


\section{Solution Approach}
\label{sec:SolutionApproach}
To solve model~\eqref{LogPosterior}, we adapt the hybrid genetic search of \citet{Gribel2019}, which has demonstrated state-of-the-art performance on the minimum-sum-of-squares clustering problem.
As summarized in Algorithm \ref{HGSearch}, the method begins with a set of~$\Pi_1$ initial solutions obtained by using the K-means algorithm starting from different centers, followed by local search.
After this initialization phase, the algorithm iteratively generates new solutions via three successive steps: crossover, mutation, and local search. Upon attaining the maximum population size $\Pi_2$, the best $\Pi_1$ solutions in terms of log-likelihood are preserved to ensure elitism and selection pressure, and the remaining solutions are discarded. The algorithm terminates after a fixed number of iterations. The remainder of this section details each operator.\\

\begin{figure}[!htbp]
  \vspace*{-0.2cm}
  \centering
    \captionsetup{justification=centering}
    \subfloat[Solution $\bm{Z}^{(1)}$]{{\includegraphics[width=4.0cm]{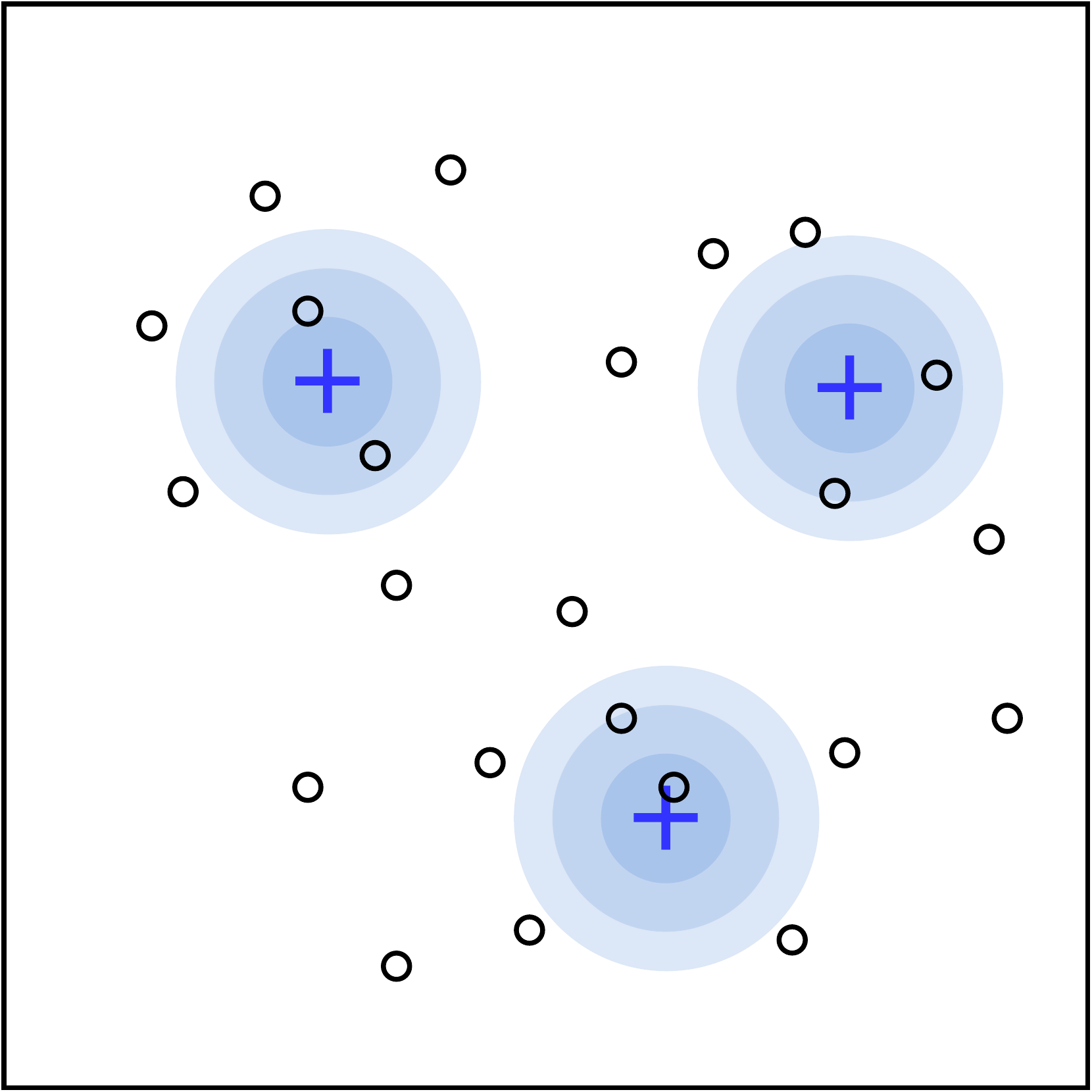}}}%
    \hspace{0.1em}
    \subfloat[Solution $\bm{Z}^{(2)}$]{{\includegraphics[width=4.0cm]{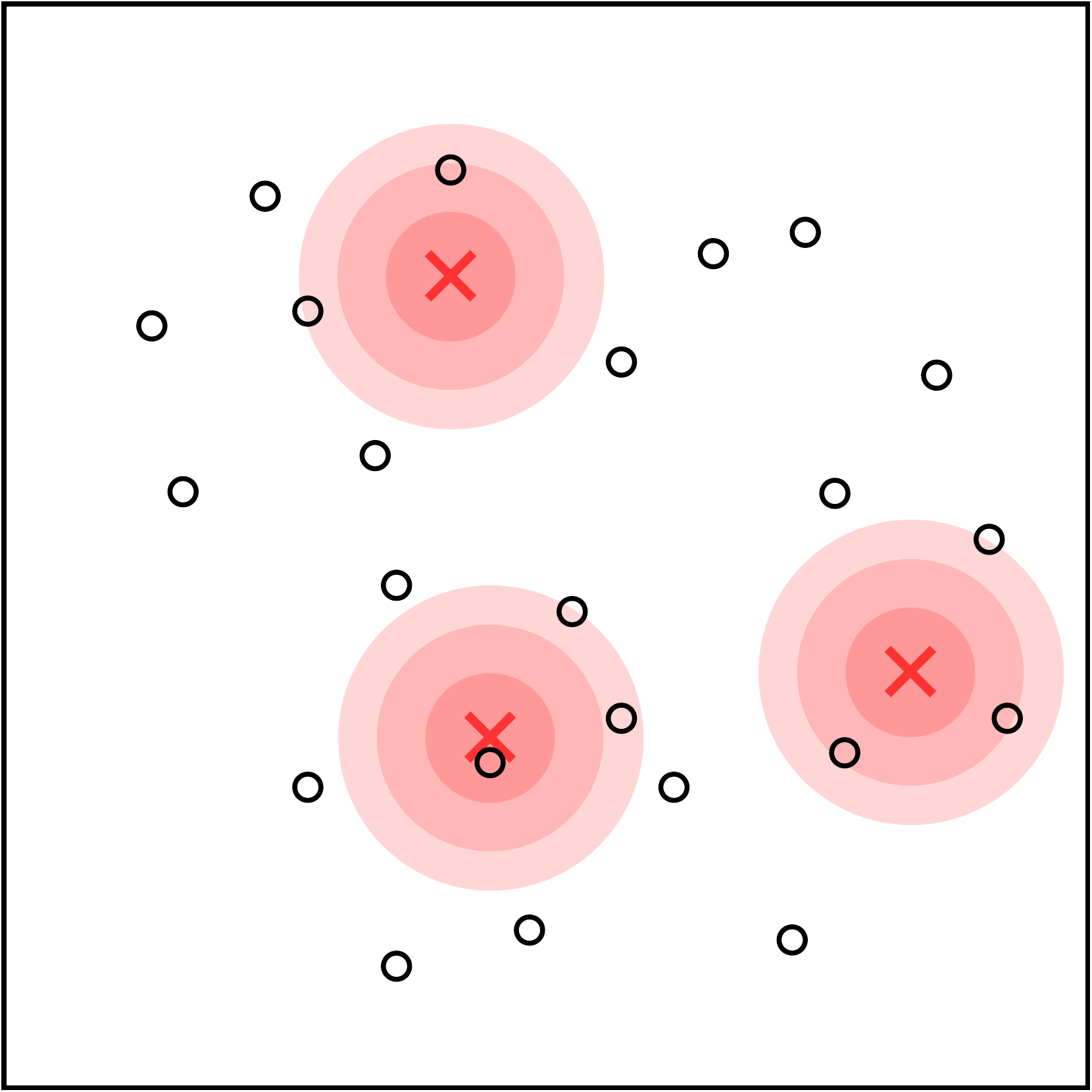}}}%
    \hspace{0.1em}
    \subfloat[Assignment and random selection]{{\includegraphics[width=4.0cm]{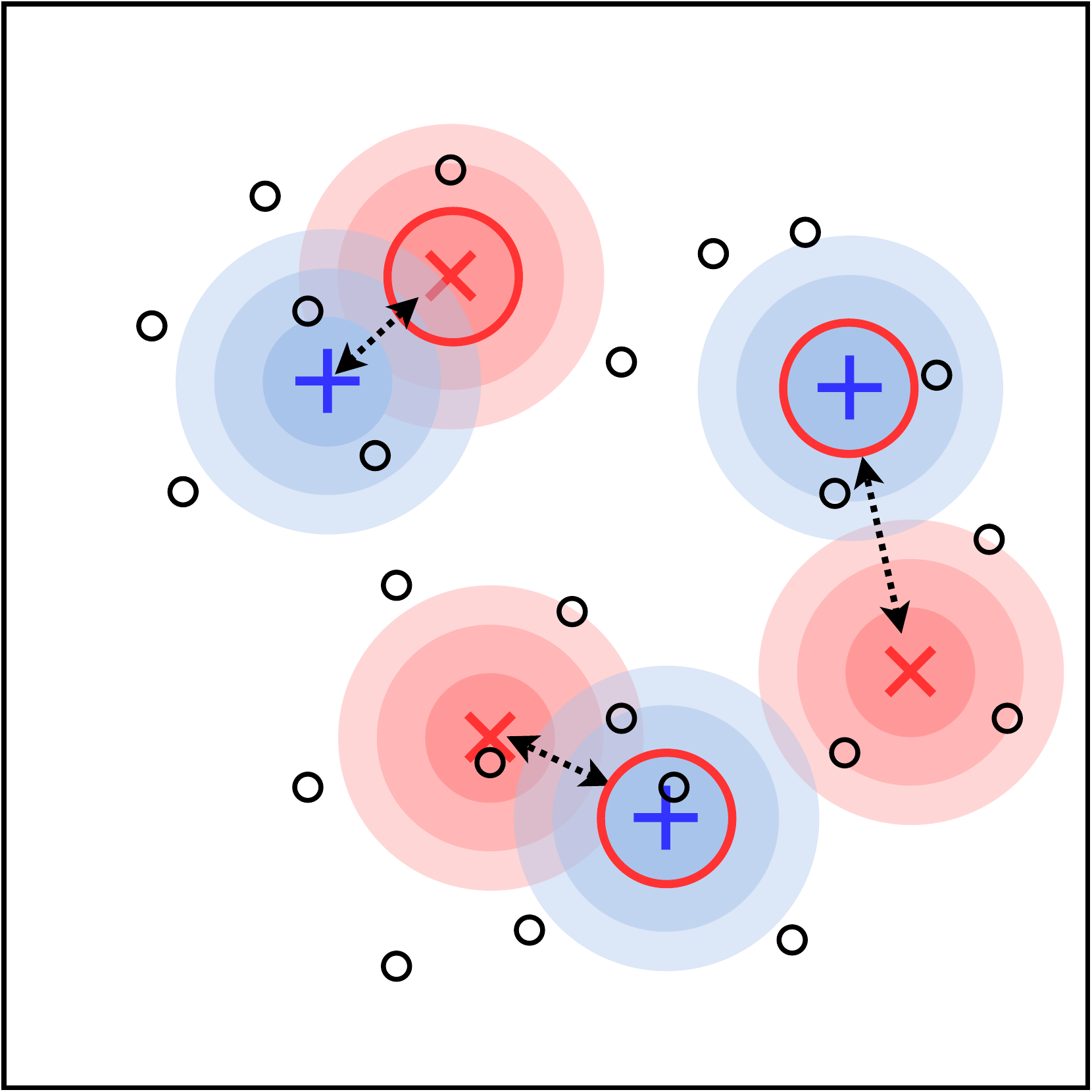}}}%
    \hspace{0.1em}
    \subfloat[Resulting solution]{{\includegraphics[width=4.0cm]{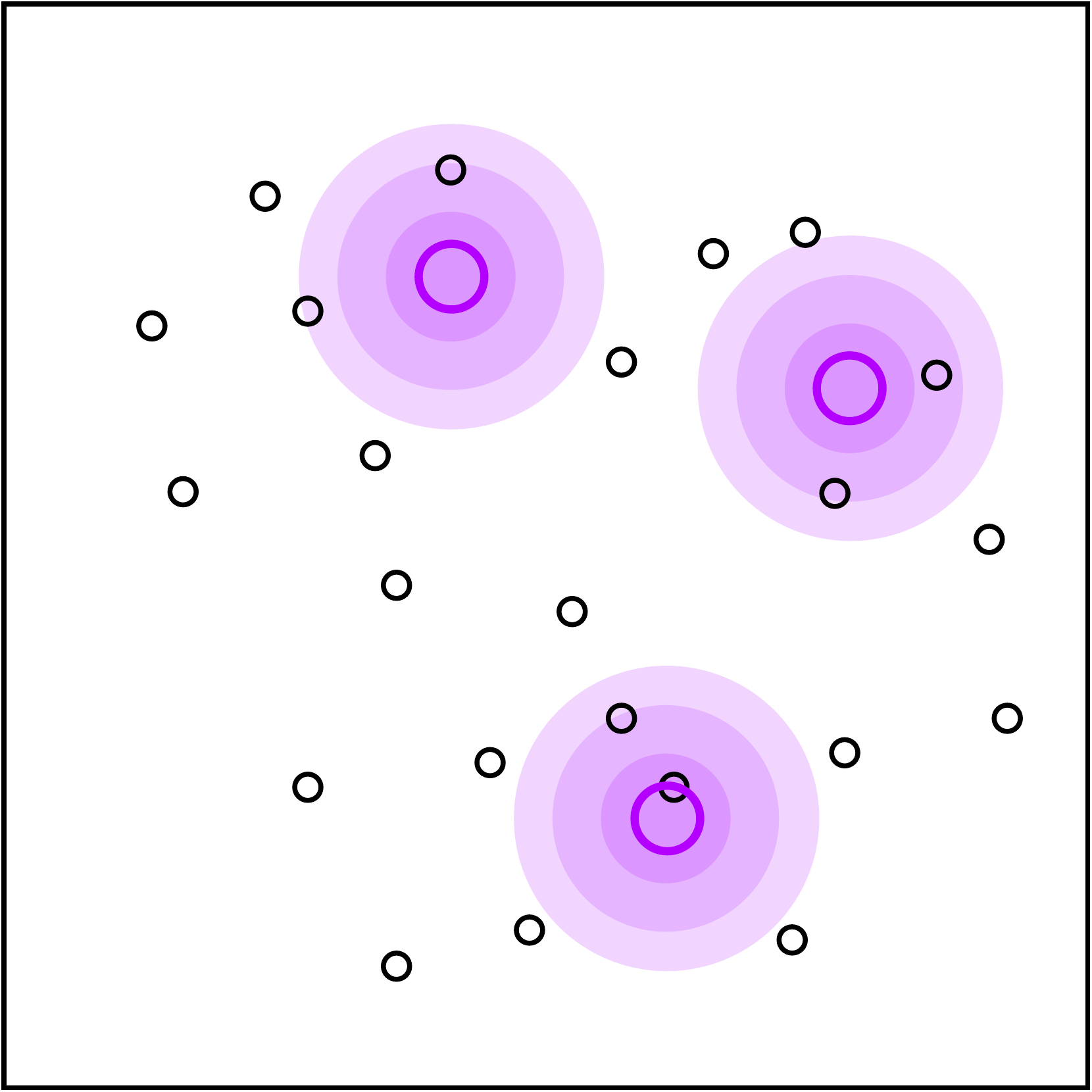}}}
    \vspace*{-0.0cm}
    \caption{Crossover based on centers' matching.}
    \label{fig:Crossover}
\end{figure}

\noindent \textbf{Crossover.} The algorithm selects two random parent solutions $\bm{Z}^{(1)}$ and $\bm{Z}^{(2)}$ in the population and applies a crossover to them to create a new solution. This operator works as follows (see Fig.~\ref{fig:Crossover}):

\begin{itemize}
\item \textbf{Step 1}. It first solves a bipartite matching problem to pair up the centers of the two solutions. Let $G = (\bm{U}, \bm{V}, \bm{E})$ be a complete bipartite graph in which the vertex set $\bm{U} = (\bm{u}_1, \dots, \bm{u}_K)$ represents the centers of solution~$\bm{Z}^{(1)}$ and $\bm{V} = (\bm{v}_1, \dots, \bm{v}_K)$ represents the centers of solution~$\bm{Z}^{(2)}$. Each edge $(\bm{u}_i, \bm{v}_j) \in \bm{E}$, for $i \in {1, \dots, K}$ and $j \in {1, \dots, K}$ represents a possible association of center $i$ from solution $\bm{Z}^{(1)}$ with center $j$ from solution $\bm{Z}^{(2)}$. The minimum-cost bipartite matching problem is then solved in  graph $G$ by considering the weights of the edges in $\bm{E}$ as the squared Euclidean distance between the vertices in $\bm{V}$ and $\bm{U}$.

\item \textbf{Step 2}. For each pair obtained in the previous step, the crossover randomly selects one of the two centers with equal probability. This  effectively  recombines the centers of both parents.

\item \textbf{Step 3}. Once the new centers are generated, each sample~$\bm{x}_i$ is assigned to the closest center in terms of Euclidean distance.\\

\end{itemize}

\noindent \textbf{Mutation.} The mutation operator follows the crossover. Its goal is to introduce randomness into the solutions and permit a broader exploration of the search space. We use a special case of the mutation scheme described in \citet{Gribel2019} in which all samples have an equal chance of being selected as the new center:
\begin{enumerate}
\item Select one center for removal with uniform probability.
\item Select a random sample and create a new center at its position.
\item Re-assign each sample to the closest center.
\end{enumerate}

\noindent \textbf{Local Search.} The solution generated by the previous steps serves as a starting point for a two-phase local search (Algorithms~\ref{FitAnnotated} and~\ref{FitUnannotated}) that iterates until converging:
\begin{enumerate}
\item The algorithm iteratively evaluates each possible relocation of an annotated sample (i.e., a sample involved in at least one pairwise annotation) to a different cluster. Each relocation is applied if it improves the likelihood (see Algorithm~\ref{FitAnnotated}).
\item Next, the unannotated samples are assigned to their closest cluster, as determined by the distance to the cluster center. The parameters of the Gaussians are then updated based on the new assignments. These two steps are iterated until convergence to a local optimum, making this step of the local search equivalent to a K-means algorithm applied to the unannotated samples (see Algorithm~\ref{FitUnannotated}).
\end{enumerate}

For notational simplicity, Algorithms~\ref{FitAnnotated} and~\ref{FitUnannotated} cover the case of log-likelihood maximization (model without priors). Still, the algorithms work analogously for the log-posterior maximization, with the priors
being updated according to Equations~\eqref{eq:RelationMust}--\eqref{eq:PriorCannot}.

\begin{algorithm}[htbp]
\caption{Hybrid-Genetic Search}
\begin{algorithmic}[1]
\STATE {\textbf{Input:} Feature data: $\bm{X}$, Adjacency matrices: $\bm{A}$, Annotated samples: $\mathcal{A}$, Unannotated samples: $\mathcal{U}$, Number of clusters: $K$, Parameters: $\Pi_1$ and $\Pi_2$}
\STATE $\mathcal{S} \leftarrow$ Set with $\Pi_1$ initial solutions
\REPEAT
        \STATE $\bm{Z}^{(1)}, \bm{Z}^{(2)} \leftarrow$ Random solutions from $\mathcal{S}$
        \STATE $\bm{Z} \leftarrow$ Crossover($\bm{Z}^{(1)}$, $\bm{Z}^{(2)}$)
        \STATE $\bm{Z'} \leftarrow$ Mutation($\bm{Z}$)
        \STATE \textbf{Algorithm~\ref{FitAnnotated}:} FitAnnotated($\bm{X}$, $\bm{A}$, $\bm{Z}', \mathcal{A}$)
        \STATE \textbf{Algorithm~\ref{FitUnannotated}:} FitUnannotated($\bm{X}$, $\bm{A}$, $\bm{Z}', \mathcal{U}$)
        \STATE Add solution $\bm{Z}'$ to $\mathcal{S}$
        \IF{$|\mathcal{S}| = \Pi_2$}
                \STATE $\mathcal{S} \leftarrow$ Select the best $\Pi_1$ solutions
        \ENDIF
\UNTIL{Maximum number of iterations is attained}
\RETURN Best solution $\bm{Z}^*$ found
\end{algorithmic}
\label{HGSearch}
\end{algorithm}

\begin{algorithm}[htbp]
\caption{FitAnnotated: Relocation of Annotated Samples}
\begin{algorithmic}[1]
\STATE {\textbf{Input:} Feature data: $\bm{X}$, Adjacency matrices: $\bm{A}$, Current solution: $\bm{Z}$, Annotated samples: $\mathcal{A}$}
\STATE Find parameters $\bm{\mu}$, $\bm{\sigma}$, $\bm{\Omega}^{+}$ and $\bm{\Omega}^{-}$ maximizing $\mathcal{L}(\bm{X}, \bm{A} | \bm{Z})$ 
\STATE Evaluate log-likelihood with the estimated parameters: $\mathcal{Q} \leftarrow \mathcal{L}(\bm{X}, \bm{A} | \bm{\mu}, \bm{\sigma}, \bm{\Omega}, \bm{Z})$
\REPEAT
        \FOR{$i \in \mathcal{A}$ and $r \in \{1,\dots,K\}$ in random order}
                \STATE Consider solution $\bm{Z}^\textsc{R}$ obtained from $\bm{Z}$ by relocating sample $i$ to cluster $r$
                \STATE Find parameters $\bm{\hat{\mu}}$, $\bm{\hat{\sigma}}$, $\bm{\hat{\Omega}}^{+}$ and $\bm{\hat{\Omega}}^{-}$ maximizing 
                $\mathcal{L}(\bm{X}, \bm{A} | \bm{Z}^\textsc{R})$ 
                \STATE Evaluate log-likelihood with the estimated parameters: $\mathcal{Q}' \leftarrow \mathcal{L}(\bm{X}, \bm{A} | \bm{\hat{\mu}}, \bm{\hat{\sigma}}, \bm{\hat{\Omega}}, \bm{Z}^\textsc{R})$
        \IF{$\mathcal{Q}' > \mathcal{Q}$}
                        \STATE Apply: $\bm{\mu} \leftarrow \bm{\hat{\mu}}$, $\bm{\sigma} \leftarrow \bm{\hat{\sigma}}$, $\bm{\Omega}^{+} \leftarrow \bm{\hat{\Omega}}^{+}$,
                        $\bm{\Omega}^{-} \leftarrow \bm{\hat{\Omega}}^{-}$,
                        \\$\bm{Z} \leftarrow \bm{Z}^\textsc{R}$, $\mathcal{Q} \leftarrow \mathcal{Q}'$
        \ENDIF
        \ENDFOR
\UNTIL{No improving relocation has been identified}
\end{algorithmic}
\label{FitAnnotated}
\end{algorithm}

\begin{algorithm}[htbp]
\caption{FitUnannotated: Assignment of Unannotated Samples}
\begin{algorithmic}[1]
\STATE {\textbf{Input:} Feature data: $\bm{X}$, Adjacency matrices: $\bm{A}$, Current solution: $\bm{Z}$, Unannotated samples:~$\mathcal{U}$}
\REPEAT
        \FOR{$i \in \mathcal{U}$}
                \STATE $y_i \leftarrow \min_{r} \left \| \bm{x}_i - \bm{\mu}_r \right \|^2$
                \STATE Update $\bm{Z}$ with the new assignment
        \ENDFOR
        \FOR{$r \in \{1, \dots, K\}$}
                \STATE $\bm{\mu}_r \leftarrow \sum_{i}^{N} z_{ir} \bm{x}_i / \sum_{i}^{N} z_{ir}$
                \STATE $\sigma_r^2 \leftarrow \sum_{i}^{N} \sum_{r}^{K} z_{ir}\left \| \bm{x}_i - \bm{\mu}_r \right \|^2 / \left ( 2D \sum_{i}^{N} z_{ir} \right )$
        \ENDFOR
\UNTIL{No change in the solution has been identified}\nobreak
\STATE Update log-likelihood with the estimated parameters and the current value of $\bm{\Omega}$:\\$\mathcal{Q} \leftarrow \mathcal{L}(\bm{X}, \bm{A} | \bm{\mu}, \bm{\sigma}, \bm{\Omega}, \bm{Z})$
\end{algorithmic}
\label{FitUnannotated}
\end{algorithm}

\section{Computational Experiments}
\label{sec:Experiments}
We conducted computational experiments to investigate two main effects. First, we analyze how the incorporation of relational information affects the performance of the method on datasets that match the ideal conditions of the model (i.e., mixtures of spherical Gaussians). We evaluate the performance of the proposed semi-supervised models as a function of the quality and  amount of the information provided and analyze the impact of incorporating prior beliefs. Second, we assess how the model performs on more challenging real data not likely to be generated from spherical Gaussian mixtures. We evaluate the extent to which the model generalizes to treat these cases and discuss some of its limitations.

All algorithms were implemented in Julia (version 1.0.5). The source code is available at \url{http://github.com/danielgribel/SSC-IPA}.

\subsection{Evaluation Metrics}
We consider three evaluation metrics in our experimental setting:  normalized mutual information (NMI)~\cite{Kvalseth1987}, an entropy-based measure to compare two partitions from the sample-group memberships; the Kullback--Leibler (KL) divergence between two Gaussians mixtures using the matching-based approximation of~\citet{Goldberger2003}; and the centroid index (CI)~\cite{Franti2014}, a discrete measure of the number of different cluster locations between two clustering solutions.
A CI of zero indicates that the given partition matches the ground-truth structure.
These metrics reflect different aspects of the solutions: NMI compares the partitions (membership variables) with  the ground-truth, the KL divergence compares the continuous Gaussian parameters, and  the CI is based on the coordinates of the solution centers.

\subsection{Performance for Mixtures of Spherical Gaussians}
In our first set of experiments, we analyze the general performance of Algorithm~\ref{HGSearch} applied to synthetic datasets that meet ideal conditions (i.e., mixtures of spherical Gaussians). 

To generate these datasets, we use overlapping mixtures in which each group has its own dispersion. More precisely, for each group $r$, we create a $D$-dimensional mean $\bm{\mu}_r$ by sampling uniformly over the range~$[-1, 1]$. For the dispersion of each group, we sample $\sigma_r^2$ uniformly from the range $[0, 5]$. 
Each data sample is then generated with probability $1/K$ from group $r$ according to the Gaussian distribution $\mathcal{N}(\bm{\mu}_r, \sigma_r^2)$. 

Finally, the edges of graphs $\bm{A}^{+}$ and $\bm{A}^{-}$ are randomly generated via Equation~\eqref{EqPoissonExpert}. We create datasets with different experts' accuracies by defining a parameter $p \in \{0.8, 0.9, 1.0\}$ and setting $p_{rr} = p$ for all $r$ and $p_{rs} = 1 - p$ for all~$r \neq s$. In all datasets, the number of samples is set to $N = 200$, the feature-space dimension is set to $D = 10$, and the number of clusters is selected from the set $K \in \{2, 4, 6\}$. For each value of $K$, we generate 50 Gaussian mixtures, leading to 150 datasets. We define the number of total annotations $m$ (including both \textit{must-links} and \textit{cannot-links}) as a proportion of the number of samples~$N$, in which~$m \in \{0,\, N/2,\, N,\, 1.5N,\, 2N,\, \dots,\, 4N\}$. This experimental setup includes 3750 cases overall, considering all 150 datasets and the possible values of~$p$ and~$m$.

\begin{table*}[!htbp]
\caption{Average NMI and KL divergence on synthetic datasets for $p = 0.8$.}
\label{tab:PerformanceSynthetic80}
\adjustbox{width=\textwidth}{
\centering
\begin{tabular}{c|cc|cc|cc|cccccc}
\hline
          & \multicolumn{6}{c|}{NMI}                                                                & \multicolumn{6}{c}{KL divergence}                                                                        \\ \hline
          & \multicolumn{2}{c|}{K = 2}  & \multicolumn{2}{c|}{K = 4}  & \multicolumn{2}{c|}{K = 6}  & \multicolumn{2}{c|}{K = 2}           & \multicolumn{2}{c|}{K = 4}           & \multicolumn{2}{c}{K = 6}  \\ \hline
Priors:   & \ding{53}           & \checkmark          & \ding{53}           & \checkmark          & \ding{53}           & \checkmark          & \ding{53}    & \multicolumn{1}{c|}{\checkmark}     & \ding{53}    & \multicolumn{1}{c|}{\checkmark}     & \ding{53}          & \checkmark          \\ \hline
$m$ = 0   & \multicolumn{2}{c|}{0.4808} & \multicolumn{2}{c|}{0.4358} & \multicolumn{2}{c|}{0.4003} & \multicolumn{2}{c|}{0.0998}          & \multicolumn{2}{c|}{0.3980}          & \multicolumn{2}{c}{0.7529} \\ \hline
$m$ = 100 & 0.5136        & 0.5160      & 0.4323        & 0.4373      & 0.3834        & 0.3878      & 0.0657 & \multicolumn{1}{c|}{0.0733} & 0.3414 & \multicolumn{1}{c|}{0.3421} & 0.7157       & 0.6671      \\
$m$ = 200 & 0.5497        & 0.5536      & 0.4319        & 0.4436      & 0.3878        & 0.3943      & 0.0528 & \multicolumn{1}{c|}{0.0520} & 0.3300 & \multicolumn{1}{c|}{0.3062} & 0.6652       & 0.6327      \\
$m$ = 300 & 0.5880        & 0.5894      & 0.4369        & 0.4468      & 0.3870        & 0.3917      & 0.0396 & \multicolumn{1}{c|}{0.0383} & 0.3188 & \multicolumn{1}{c|}{0.3089} & 0.6093       & 0.6121      \\
$m$ = 400 & 0.6676        & 0.6549      & 0.4687        & 0.4697      & 0.3976        & 0.4064      & 0.0285 & \multicolumn{1}{c|}{0.0327} & 0.2614 & \multicolumn{1}{c|}{0.2563} & 0.5903       & 0.5790      \\
$m$ = 500 & 0.7260        & 0.7210      & 0.4895        & 0.4865      & 0.3907        & 0.4063      & 0.0154 & \multicolumn{1}{c|}{0.0152} & 0.2380 & \multicolumn{1}{c|}{0.2365} & 0.6013       & 0.5550      \\
$m$ = 600 & 0.7974        & 0.7978      & 0.4994        & 0.5030      & 0.4006        & 0.4201      & 0.0075 & \multicolumn{1}{c|}{0.0077} & 0.2206 & \multicolumn{1}{c|}{0.2167} & 0.5637       & 0.5276      \\
$m$ = 700 & 0.8328        & 0.8346      & 0.5296        & 0.5375      & 0.4114        & 0.4216      & 0.0060 & \multicolumn{1}{c|}{0.0054} & 0.2092 & \multicolumn{1}{c|}{0.1757} & 0.5563       & 0.5413      \\
$m$ = 800 & 0.8616        & 0.8623      & 0.5369        & 0.5479      & 0.4181        & 0.4322      & 0.0037 & \multicolumn{1}{c|}{0.0036} & 0.2102 & \multicolumn{1}{c|}{0.1679} & 0.5404       & 0.4985      \\ \hline
\end{tabular}}
\end{table*}

\begin{table*}[!htbp]
\caption{Average NMI and KL divergence on synthetic datasets for $p = 0.9$.}
\label{tab:PerformanceSynthetic90}
\adjustbox{width=\textwidth}{
\centering
\begin{tabular}{c|cc|cc|cc|cccccc}
\hline
          & \multicolumn{6}{c|}{NMI}                                                                & \multicolumn{6}{c}{KL divergence}                                                                              \\ \hline
          & \multicolumn{2}{c|}{K = 2}  & \multicolumn{2}{c|}{K = 4}  & \multicolumn{2}{c|}{K = 6}  & \multicolumn{2}{c|}{K = 2}              & \multicolumn{2}{c|}{K = 4}              & \multicolumn{2}{c}{K = 6}  \\ \hline
Priors:   & \ding{53}           & \checkmark          & \ding{53}           & \checkmark          & \ding{53}           & \checkmark          & \ding{53}    & \multicolumn{1}{c|}{\checkmark}     & \ding{53}    & \multicolumn{1}{c|}{\checkmark}     & \ding{53}          & \checkmark          \\ \hline
$m$ = 0   & \multicolumn{2}{c|}{0.4808} & \multicolumn{2}{c|}{0.4358} & \multicolumn{2}{c|}{0.4003} & \multicolumn{2}{c|}{0.0998}             & \multicolumn{2}{c|}{0.3980}             & \multicolumn{2}{c}{0.7529} \\ \hline
$m$ = 100 & 0.5461       & 0.5483       & 0.4513       & 0.4607       & 0.3930       & 0.4034       & 0.0541  & \multicolumn{1}{c|}{0.0537}   & 0.3291  & \multicolumn{1}{c|}{0.2906}   & 0.6637       & 0.6433      \\
$m$ = 200 & 0.6515       & 0.6601       & 0.4672       & 0.4840       & 0.4048       & 0.4089       & 0.0304  & \multicolumn{1}{c|}{0.0263}   & 0.2951  & \multicolumn{1}{c|}{0.2702}   & 0.6311       & 0.6043      \\
$m$ = 300 & 0.7546       & 0.7618       & 0.4998       & 0.5140       & 0.4046       & 0.4118       & 0.0117  & \multicolumn{1}{c|}{0.0095}   & 0.2531  & \multicolumn{1}{c|}{0.2157}   & 0.6225       & 0.5669      \\
$m$ = 400 & 0.8603       & 0.8678       & 0.5369       & 0.5529       & 0.4248       & 0.4485       & 0.0049  & \multicolumn{1}{c|}{0.0041}   & 0.2155  & \multicolumn{1}{c|}{0.1785}   & 0.5545       & 0.4835      \\
$m$ = 500 & 0.9045       & 0.9017       & 0.5998       & 0.6085       & 0.4385       & 0.4562       & 0.0020  & \multicolumn{1}{c|}{0.0020}   & 0.1635  & \multicolumn{1}{c|}{0.1362}   & 0.5284       & 0.4684      \\
$m$ = 600 & 0.9381       & 0.9387       & 0.6416       & 0.6694       & 0.4731       & 0.4908       & 0.0012  & \multicolumn{1}{c|}{0.0012}   & 0.1314  & \multicolumn{1}{c|}{0.1004}   & 0.4753       & 0.4098      \\
$m$ = 700 & 0.9664       & 0.9659       & 0.7107       & 0.7302       & 0.4768       & 0.5035       & 0.0007  & \multicolumn{1}{c|}{0.0007}   & 0.0825  & \multicolumn{1}{c|}{0.0617}   & 0.4823       & 0.4022      \\
$m$ = 800 & 0.9722       & 0.9728       & 0.7608       & 0.7831       & 0.4991       & 0.5274       & 0.0006  & \multicolumn{1}{c|}{0.0005}   & 0.0562  & \multicolumn{1}{c|}{0.0492}   & 0.4518       & 0.3476      \\ \hline
\end{tabular}}
\end{table*}

\begin{table*}[!htbp]
\caption{Average NMI and KL divergence on synthetic datasets for $p = 1.0$.}
\label{tab:PerformanceSynthetic100}
\adjustbox{width=\textwidth}{
\centering
\begin{tabular}{c|cc|cc|cc|cccccc}
\hline
          & \multicolumn{6}{c|}{NMI}                                                                & \multicolumn{6}{c}{KL divergence}                                                                              \\ \hline
          & \multicolumn{2}{c|}{K = 2}  & \multicolumn{2}{c|}{K = 4}  & \multicolumn{2}{c|}{K = 6}  & \multicolumn{2}{c|}{K = 2}              & \multicolumn{2}{c|}{K = 4}              & \multicolumn{2}{c}{K = 6}  \\ \hline
Priors:   & \ding{53}           & \checkmark          & \ding{53}           & \checkmark          & \ding{53}           & \checkmark          & \ding{53}    & \multicolumn{1}{c|}{\checkmark}     & \ding{53}    & \multicolumn{1}{c|}{\checkmark}     & \ding{53}          & \checkmark          \\ \hline
$m$ = 0   & \multicolumn{2}{c|}{0.4808} & \multicolumn{2}{c|}{0.4358} & \multicolumn{2}{c|}{0.4003} & \multicolumn{2}{c|}{0.0998}             & \multicolumn{2}{c|}{0.3980}             & \multicolumn{2}{c}{0.7529} \\ \hline
$m$ = 100 & 0.6444       & 0.5559       & 0.4683       & 0.4706       & 0.4019       & 0.4156       & 0.0293  & \multicolumn{1}{c|}{0.0401}   & 0.3213  & \multicolumn{1}{c|}{0.2713}   & 0.6843       & 0.6002      \\
$m$ = 200 & 0.8140       & 0.7578       & 0.5195       & 0.5218       & 0.4228       & 0.4551       & 0.0078  & \multicolumn{1}{c|}{0.0086}   & 0.2741  & \multicolumn{1}{c|}{0.1939}   & 0.6060       & 0.4648      \\
$m$ = 300 & 0.9402       & 0.9311       & 0.6128       & 0.6016       & 0.4642       & 0.4820       & 0.0012  & \multicolumn{1}{c|}{0.0015}   & 0.1703  & \multicolumn{1}{c|}{0.1145}   & 0.5468       & 0.3874      \\
$m$ = 400 & 0.9746       & 0.9746       & 0.7075       & 0.7145       & 0.4926       & 0.5270       & 0.0004  & \multicolumn{1}{c|}{0.0004}   & 0.1039  & \multicolumn{1}{c|}{0.0616}   & 0.4818       & 0.3122      \\
$m$ = 500 & 0.9936       & 0.9936       & 0.8089       & 0.8289       & 0.5409       & 0.5891       & 0.0001  & \multicolumn{1}{c|}{0.0001}   & 0.0602  & \multicolumn{1}{c|}{0.0238}   & 0.4167       & 0.2388      \\
$m$ = 600 & 0.9976       & 0.9976       & 0.8830       & 0.9130       & 0.6120       & 0.6508       & 0.0000  & \multicolumn{1}{c|}{0.0000}   & 0.0277  & \multicolumn{1}{c|}{0.0086}   & 0.3182       & 0.1666      \\
$m$ = 700 & 0.9976       & 0.9976       & 0.9375       & 0.9490       & 0.6602       & 0.7322       & 0.0001  & \multicolumn{1}{c|}{0.0001}   & 0.0110  & \multicolumn{1}{c|}{0.0056}   & 0.2838       & 0.1110      \\
$m$ = 800 & 1.0000       & 1.0000       & 0.9678       & 0.9749       & 0.7509       & 0.8038       & 0.0000  & \multicolumn{1}{c|}{0.0000}   & 0.0065  & \multicolumn{1}{c|}{0.0026}   & 0.1848       & 0.0740      \\ \hline
\end{tabular}}
\end{table*}

\begin{figure*}[!htbp]
    \centering
    \vspace*{-0.5cm}
    \captionsetup{justification=centering}
    \subfloat[$p = 0.8$]{{\includegraphics[width=5.8cm]{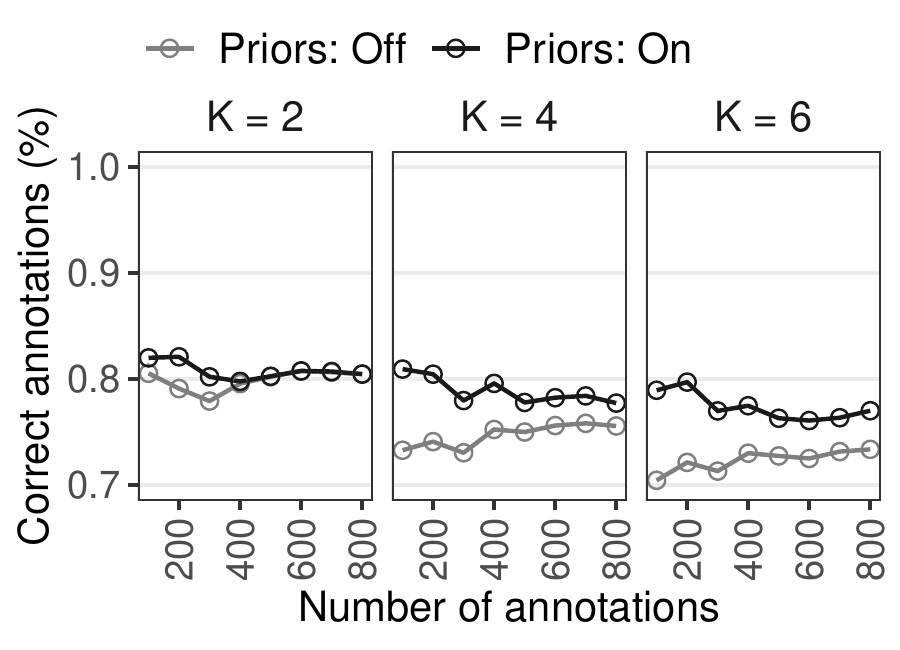}}}%
    \hspace{0.5em}
    \subfloat[$p = 0.9$]{{\includegraphics[width=5.8cm]{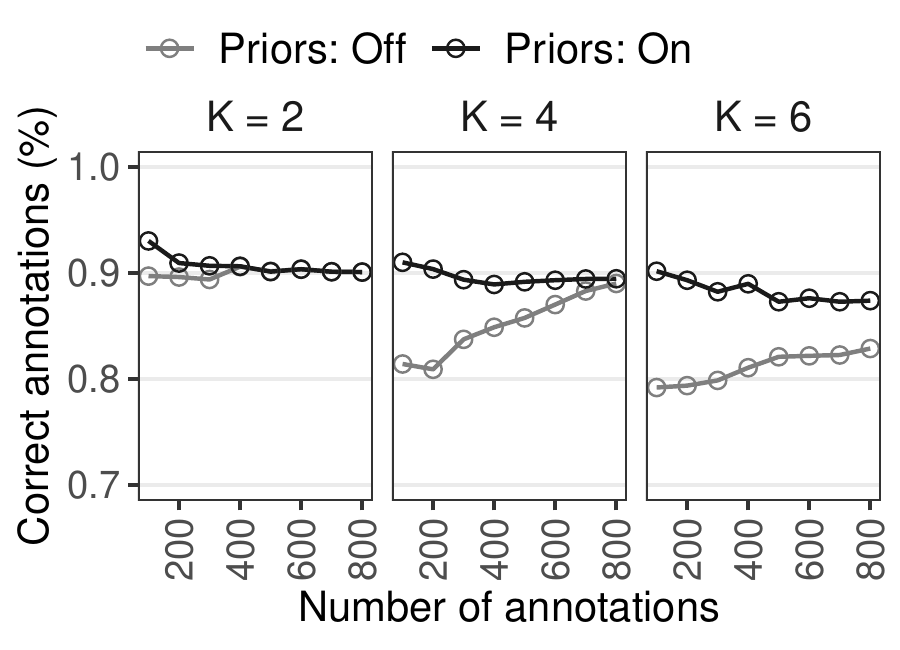}}}%
    \hspace{0.5em}
    \subfloat[$p = 1.0$]{{\includegraphics[width=5.8cm]{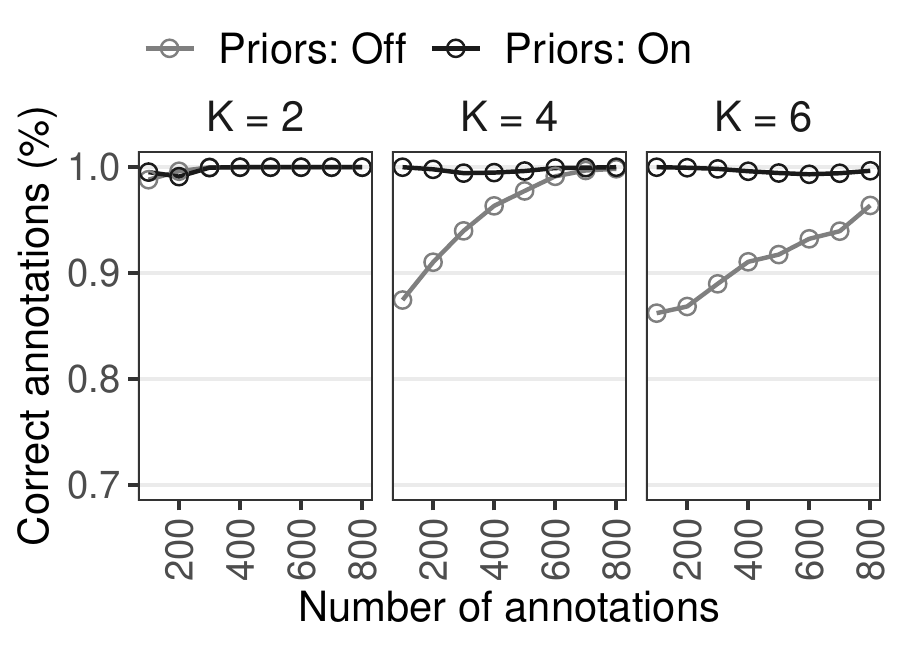}}}%
    \caption{Percentage of correct annotations given by the two proposed models in synthetic datasets.}
    \label{fig:ViolPerformance}
\end{figure*}

Tables~\ref{tab:PerformanceSynthetic80}--\ref{tab:PerformanceSynthetic100} report the performance of both proposed models for $p \in \{0.8, 0.9, 1.0\}$. All results correspond to the best log-likelihood solution found after 50 repetitions of Algorithm~\ref{HGSearch}.

Table~\ref{tab:PerformanceSynthetic80} reports the NMI and KL divergence performance for $p = 0.8$ when we expect a mistake rate of $20\%$ for the experts. For $K = 2$, we observe that the pairwise annotations have a positive impact on clustering performance, even with a small amount of information. For 100 annotations, the use of pairwise information leads to an average NMI of approximately $0.51$, against $0.48$ for the unsupervised model ($m = 0$). For datasets with four clusters,  a significant performance enhancement occurs only for $m \geq 400$. With six clusters, the incorporation of pairwise annotations has less impact. Moreover, the NMI slightly decreases for small values of $m$ despite the improved KL divergence. Finally, the inclusion of priors does not lead to large performance differences in this setting. The most visible impact occurs when $K = 4$ and $K = 6$, but only when $m$ is sufficiently~large.

Table~\ref{tab:PerformanceSynthetic90} presents the same set of experiments for $p~=~0.9$. Since the annotation accuracy is higher than in the previous case, the resulting graphs are more structured. Consequently, semi-supervision translates into a larger gain of performance over the unsupervised model. For $K~=~2$ and $m = 100$, the semi-supervised models present an average NMI of approximately $0.55$. In the case with two clusters, the proposed models achieve a near-perfect recovery when $m$ is large.
Finally, the use of prior information regarding the experts' accuracy had a more significant impact than in the previous case with $p = 0.8$.

As expected and seen in Table~\ref{tab:PerformanceSynthetic100}, the difference in performance between the semi-supervised and unsupervised approaches becomes evident in a regime with perfect annotations $p = 1.0$. With $K=2$ and $m = 100$ annotations, we obtain an average NMI of more than $0.64$ without priors. As the number of annotations grows, the semi-supervised solution converges toward the ground-truth, effectively attaining it when $m = 800$ and $K = 2$. Still,  when $p = 1.0$, the model with priors suffers from numerical instability because~$f_{\text{IN}}^{-}$ and~$f_{\text{OUT}}^{+}$ drop to zero. To circumvent this issue, we use~$p = 1 - 10^{-6}$ as an approximation. Despite this adjustment, the penalties represented by $\bm{\lambda}$ may still remain quite large such that, for small values of $K$ and sparse graphs with many unannotated samples, the priors tend to dominate the other terms in the objective function. This diminishes the impact of the Gaussians terms in the objective and leads to more frequent misallocations of unannotated samples.
We therefore recommend using the formulation without priors in these circumstances or even using simple constraints when the experts' annotations are perfect. In the other circumstances, the model with priors generally performs better.

Fig.~\ref{fig:ViolPerformance} presents the average percentage of correct annotations according to the partitions obtained with the two models. When we incorporate the prior beliefs, this quantity becomes close to the real number of correct annotations for $K = 4$ and $K = 6$. Conversely, the model without priors requires more information to approximate the real number of mistakes even when the experts' accuracy is high ($p = 0.9$ and $p = 1.0$).
This behavior stems from the fact that ordinary SBMs can recover any connectivity pattern, which may be an issue in sparse graphs with little structure~\citep{Gribel2020}.

\begin{figure*}[!t]
    \centering
    \vspace*{-0.1cm}
    \captionsetup{justification=centering}
    {\includegraphics[width=\textwidth]{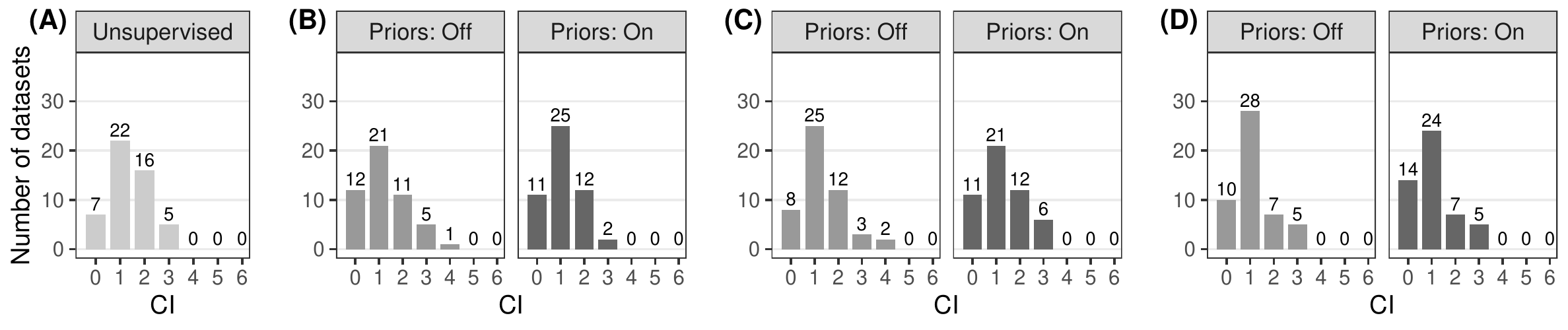}}%
    \vspace*{-0.4cm}\caption{CI of 50 Gaussian mixtures for (A) $m = 0$, (B) $m = 200$, (C)\ $m = 400$, and (D) $m = 600$, with $p = 0.8$.}
    \label{fig:CiHistogram-6-80}
\end{figure*}

\begin{figure*}[!t]
    \centering
    \vspace*{-0.1cm}
    \captionsetup{justification=centering}
    {\includegraphics[width=\textwidth]{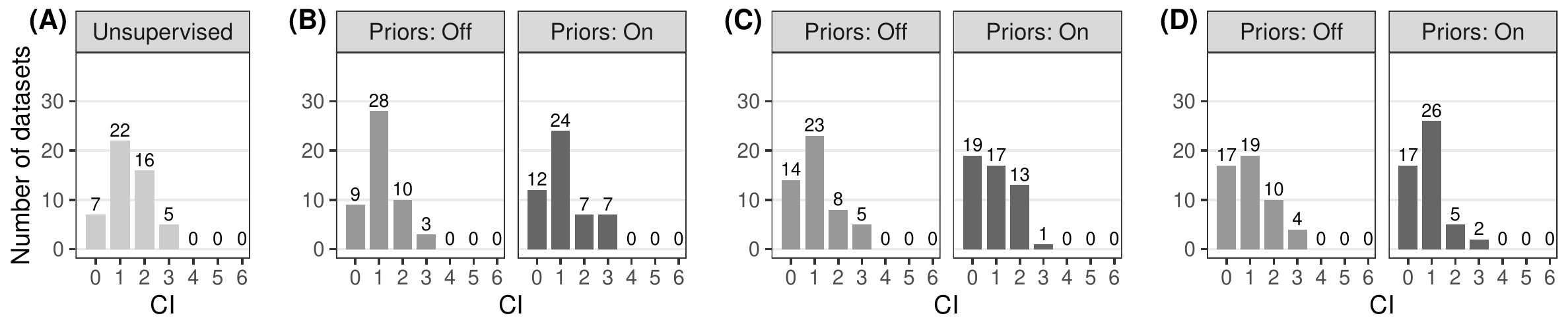}}%
    \vspace*{-0.4cm}\caption{CI of 50 Gaussian mixtures for (A) $m = 0$, (B) $m = 200$, (C) $m = 400$, and (D)\ $m = 600$, with $p = 0.9$.}
    \label{fig:CiHistogram-6-90}
\end{figure*}

\begin{figure*}[!t]
    \centering
    \vspace*{-0.1cm}
    \captionsetup{justification=centering}
    {\includegraphics[width=\textwidth]{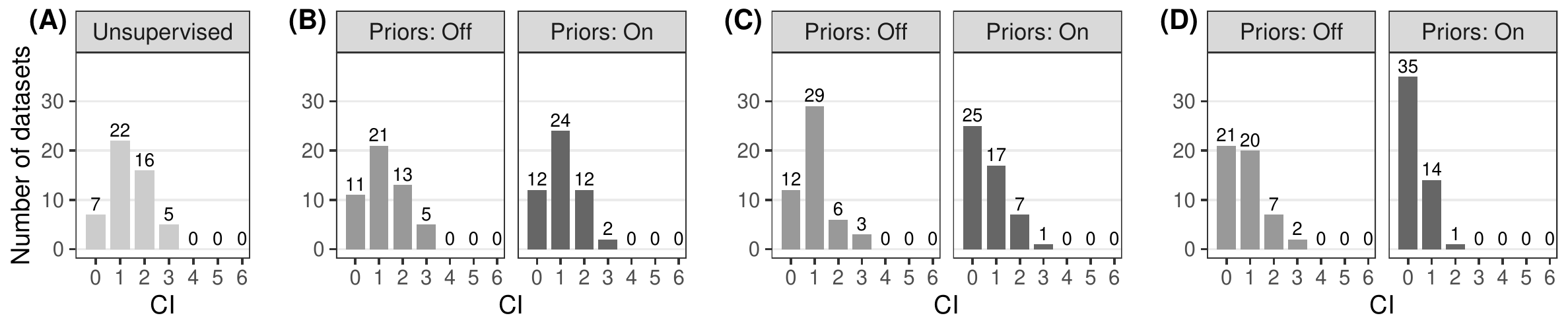}}%
    \vspace*{-0.4cm}\caption{CI of 50 Gaussian mixtures for (A) $m = 0$, (B) $m = 200$, (C) $m = 400$, and (D) $m = 600$, with $p = 1.0$.}
    \label{fig:CiHistogram-6-100}
\end{figure*}

Finally, Figs.~\ref{fig:CiHistogram-6-80}--\ref{fig:CiHistogram-6-100} compare the CI obtained with the two proposed models and the unsupervised model for $K = 6$ and different values of $m$. In Fig.~\ref{fig:CiHistogram-6-80}, for $p = 0.8$, no significant difference appears between the three models,
although the semi-supervised models present more datasets with CI~=~0 (same ground-truth structure) and CI~=~1 (one center diverging from the ground-truth center locations). Fig.~\ref{fig:CiHistogram-6-90} compares the CI when $p = 0.9$. For 200 annotations, 37 out of 50 datasets have $\text{CI} = 0$ or $\text{CI}=1$ without prior information, whereas 36 cases are reported with priors. The unsupervised model, however, presents only 29 datasets with   $\text{CI}= 0$ or 1. Finally, Fig.~\ref{fig:CiHistogram-6-100} shows the CI distribution with perfect annotation accuracy. In this case, differences between the two proposed models are more significant, notably when more information is provided.

\subsection{Performance on Real-World Benchmarks}
This section considers datasets that are  assumed not to be generated by spherical Gaussian distributions. The goal is to show whether the introduction of pairwise information leads to partitions with a different structure from those obtained with unsupervised clustering in challenging datasets that do not fit the  original assumptions of the model. For this analysis, we consider eight real datasets from the UCI machine learning repository \citep{Dua2019} with continuous multi-feature data and available ground-truth information. Table~\ref{UCI-Summary} summarizes these datasets in terms of size and number of clusters.

\vspace*{-0.1cm}
\begin{table}[!htbp]
\caption{UCI datasets.}
\renewcommand{\arraystretch}{1.2}
\label{UCI-Summary}
\centering
\vspace*{-0.2cm}
\begin{tabular}{lccc}
\hline
Dataset       & N    & D  & K \\ \hline
Diabetes      & 145  & 5  & 3 \\
Iris          & 150  & 4  & 3 \\
Wine          & 178  & 13 & 3 \\
Thyroid       & 215  & 5  & 3 \\
Vertebral     & 310  & 6  & 3 \\
E. coli       & 336  & 7  & 8 \\
Breast-Cancer & 683  & 9  & 2 \\
Pendigits-389 & 2157 & 16 & 3 \\
\hline
\end{tabular}
\end{table}

\begin{figure*}[!t]
    \centering
    \vspace*{-0.1cm}
    \captionsetup{justification=centering}
    {\includegraphics[width=\textwidth]{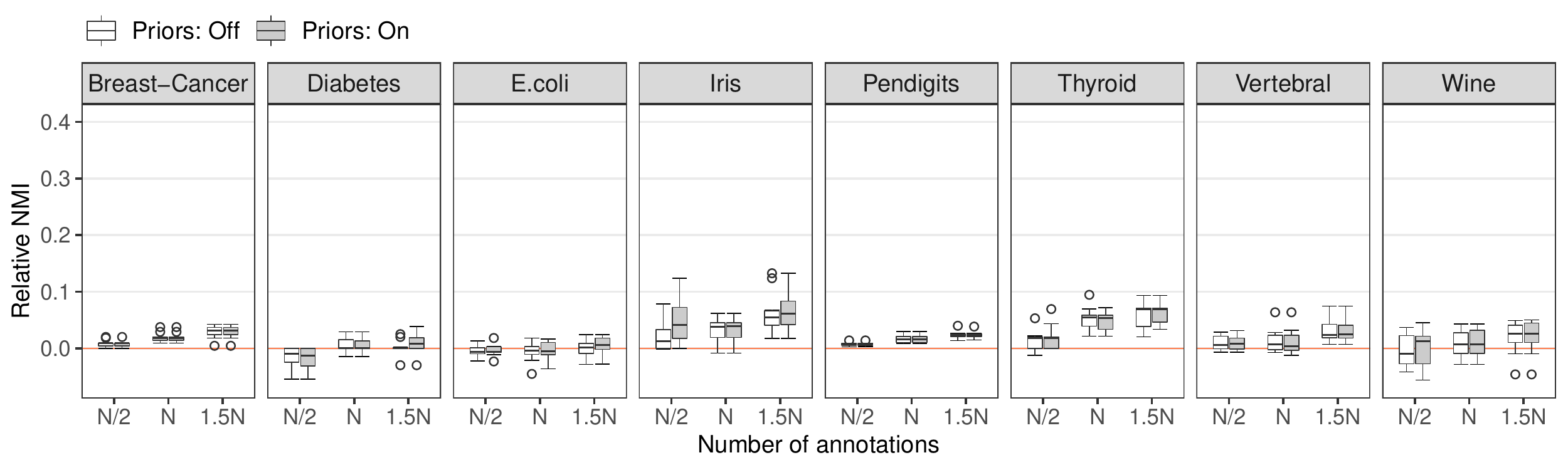}}%
    \vspace*{-0.3cm}\caption{Relative NMI between  proposed models and  unsupervised model for UCI datasets ($p = 0.8$).}
    \label{fig:BoxplotUCI-80}
\end{figure*}

\begin{figure*}[!t]
    \centering
    \vspace*{-0.1cm}
    \captionsetup{justification=centering}
    {\includegraphics[width=\textwidth]{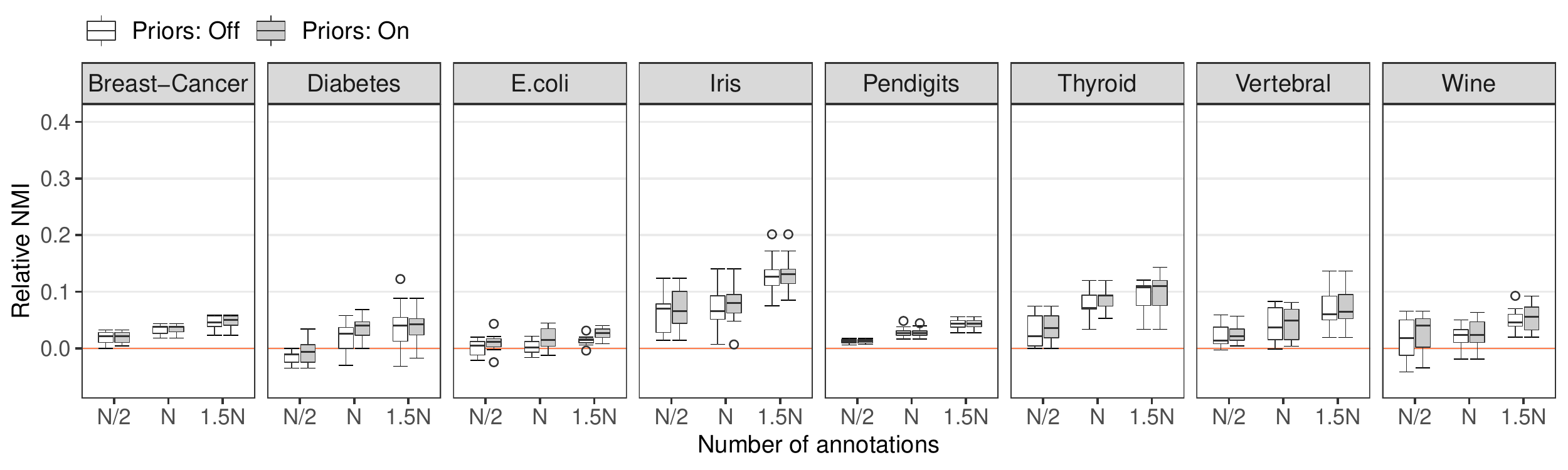}}%
    \vspace*{-0.3cm}\caption{Relative NMI between  proposed models and  unsupervised model for UCI datasets ($p = 0.9$).}
    \label{fig:BoxplotUCI-90}
\end{figure*}

\begin{figure*}[!t]
    \centering
    \vspace*{-0.1cm}
    \captionsetup{justification=centering}
    {\includegraphics[width=\textwidth]{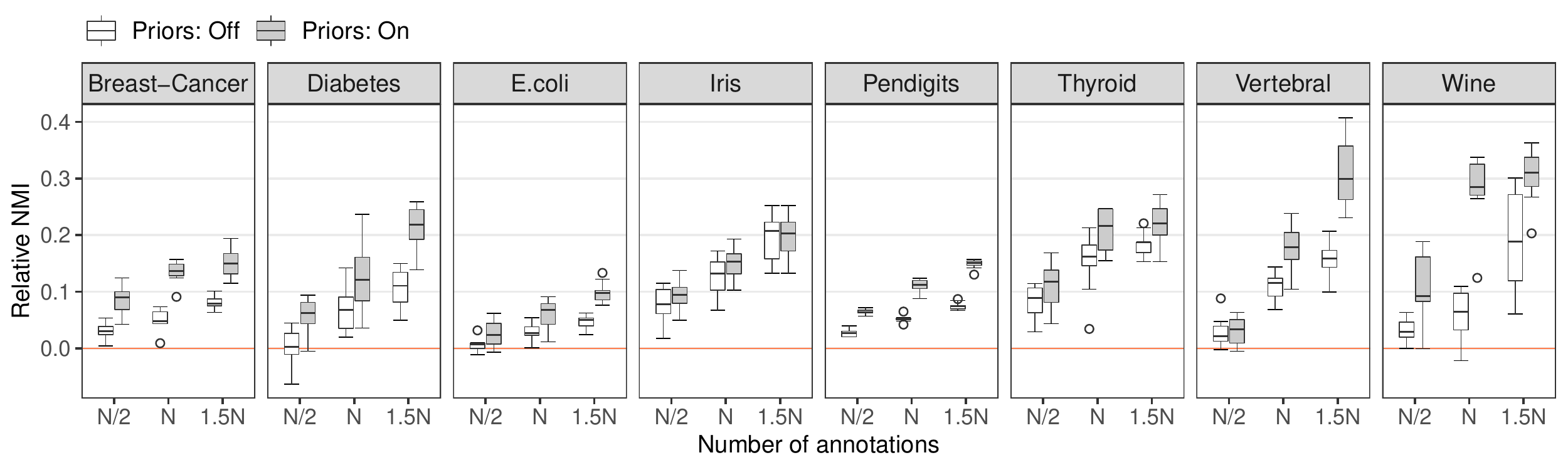}}%
    \vspace*{-0.3cm}\caption{Relative NMI between  proposed models and  unsupervised model for UCI datasets ($p = 1.0$).}
    \label{fig:BoxplotUCI-100}
\end{figure*}

Figs.~\ref{fig:BoxplotUCI-80} to~\ref{fig:BoxplotUCI-100} show the performance of the two models in terms of NMI for $p \in \{0.8,\, 0.9, \,1.0\}$ and $m \in \{N/2, \,N,\, 1.5N\}$. 
For each combination of a dataset and values of $p$ and $m$, we generate ten different graphs.
We run 50 repetitions of Algorithm~\ref{HGSearch} on each case and register the NMI for the solution with the best log-likelihood.
Then, we measure the difference of (i.e., relative) NMI between each of the proposed semi-supervised models and the baseline model without supervision and represent those values as boxplots, in which the whiskers extend to 1.5 times the interquartile range.

Fig.~\ref{fig:BoxplotUCI-80} presents the relative NMI 
for $p = 0.8$, i.e., considering annotations that are quite inaccurate. In three out of eight datasets, the NMI improves when pairwise information is considered. 
For the remaining datasets, no significant improvement occurs. Additionally, we did not observe significant differences between the two proposed models (with or without priors) for $p = 0.8$. In general, prior knowledge led to the same solutions that are obtained without priors. 

Fig.~\ref{fig:BoxplotUCI-90} reports the relative NMI for graphs with only~$10\%$ annotation errors ($p = 0.9$). In these conditions, the NMI obtained in the datasets \textit{Diabetes}, \textit{E. coli}, and \textit{Wine} also increase visibly upon adding pairwise information. Although the two proposed models perform quite similarly, in all test instances, the model with priors performs the same or better than the model without priors.

In Fig.~\ref{fig:BoxplotUCI-100}, with perfect annotations, the median relative NMI is positive in all datasets and for all values of $m$. We consider again $p = 1 - 10^{-6}$ for the priors estimation. A~notable difference now appears between the two proposed models. The most expressive difference appears for $m = 1.5N$ (average NMI of $0.8828$ with prior information versus $0.8074$ without priors).
The results reveal that, given trusted supervision,  attaching prior beliefs significantly boosts performance even if the available supervision is limited and the datasets do not fit the original assumptions.\\

\begin{figure}[!htbp]
    \centering
    \vspace*{-0.0cm}
    \captionsetup{justification=centering}
    \subfloat[Unsupervised clustering]{{\includegraphics[width=0.39\textwidth]{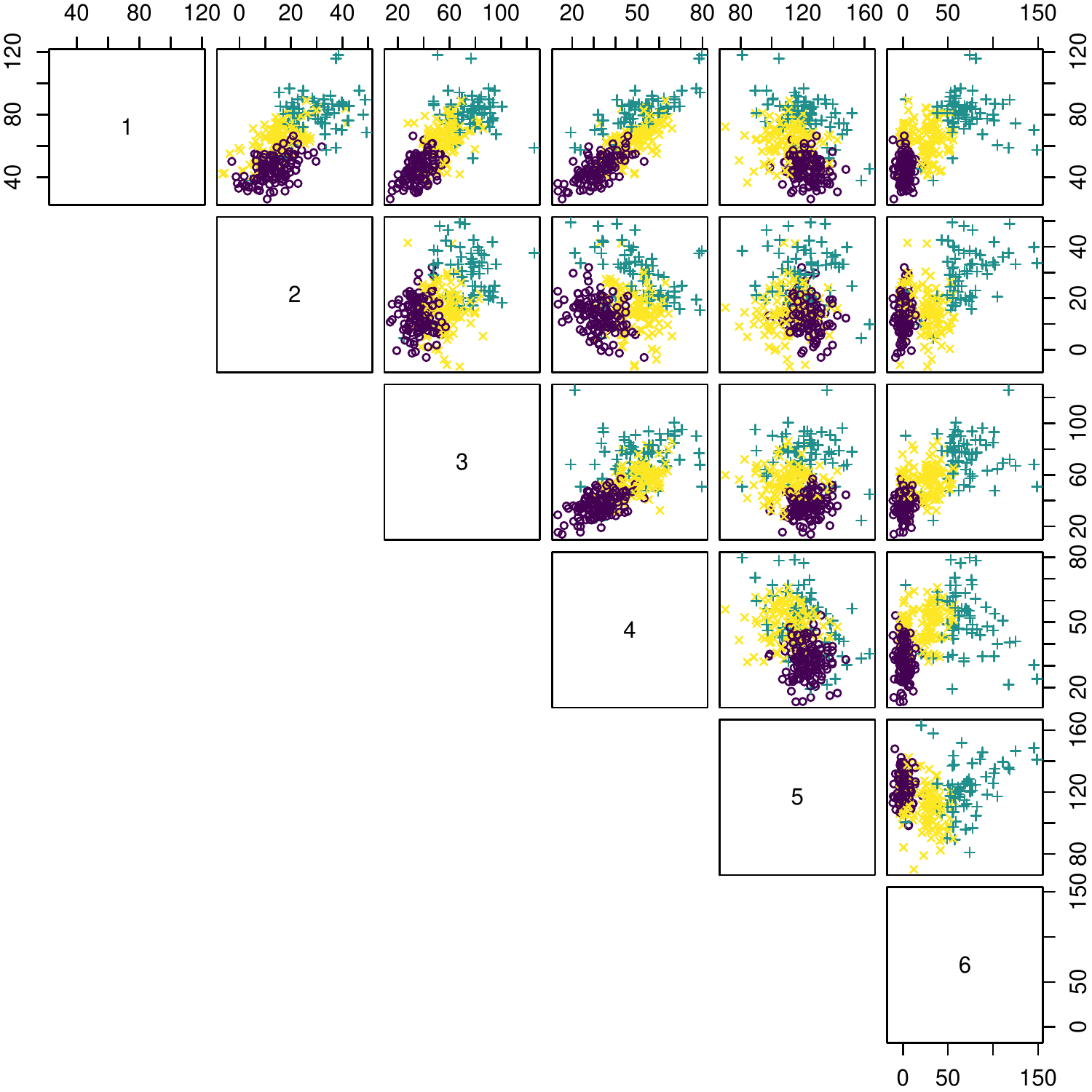}}}%
    \hspace{0.1em}
    \subfloat[Pairwise supervision $(m = N)$]{{\includegraphics[width=0.39\textwidth]{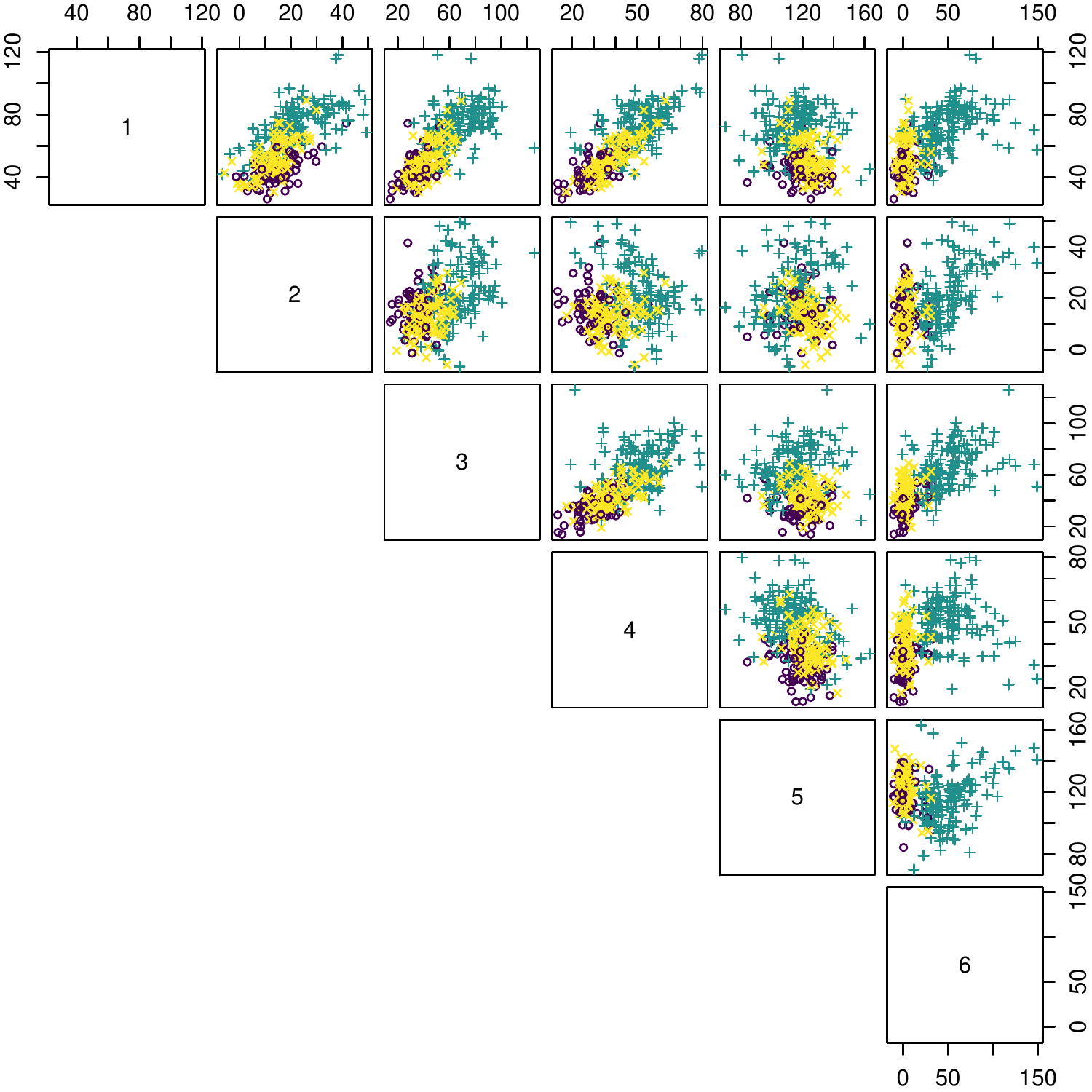}}}%
    \hspace{0.1em}
    \subfloat[Ground-truth solution]{{\includegraphics[width=0.39\textwidth]{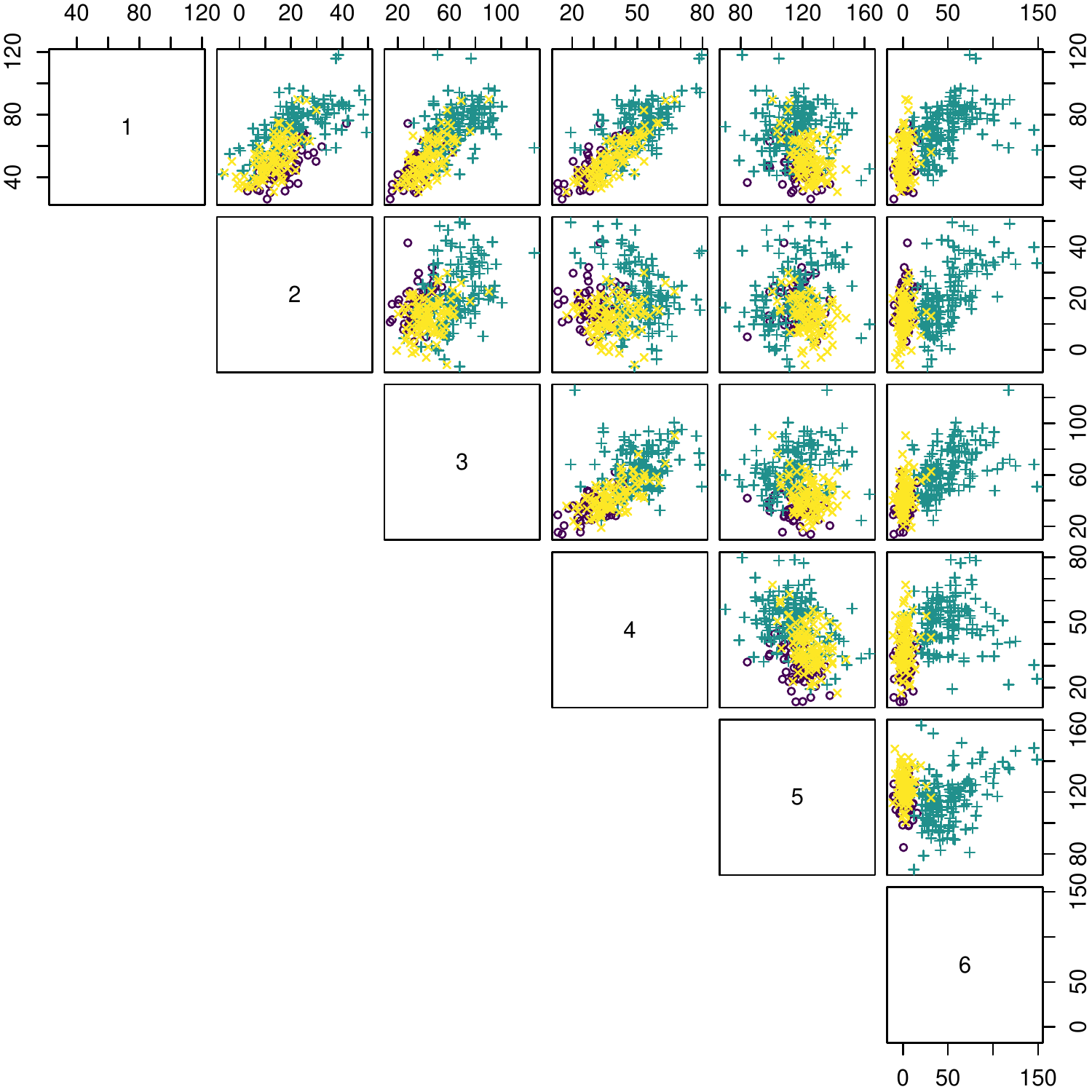}}}%
    \caption{Different clustering structures found in the \textit{Vertebral} dataset.}
    \label{fig:Vertebral}
\end{figure}

\noindent \textbf{Vertebral column dataset.}  Fig.~\ref{fig:Vertebral} presents the solutions obtained for the \textit{Vertebral} dataset \cite{Dua2019} with unsupervised clustering and the semi-supervised model with priors, along with the ground-truth solution. The \textit{Vertebral} dataset contains six biomechanical measures used to classify orthopedic patients into three classes: normal, disk hernia, and \textit{spondilolysthesis}. Each diagonal in the figure represents one feature of the dataset, and each upper square presents the feature values in pairs. We consider $m = N = 310$ pairwise annotations with no errors. In these conditions, the unsupervised model obtains CI = $1$, whereas the use of pairwise annotations leads to a CI of zero. Without side information, unsupervised clustering naturally tends to retrieve partitions with separable clusters because it relies only on the available features. The introduction of pairwise information can reveal hidden structures, especially if some clusters  significantly overlap, which is the case for the \textit{Vertebral} dataset. The yellow (``$\times$'' crosses) and purple (circles) clusters considerably overlap, and the unsupervised formulation does not capture this characteristic. Adding pairwise annotations can guide the clustering process toward solutions that differ structurally from those obtained with an unsupervised model. Beyond repairing the membership of annotated samples originally misallocated by unsupervised methods, the semi-supervision also reveals partitions with markedly distinct structures.

\section{Conclusion}
Side information in the form of pairwise annotations can be a powerful tool to improve clustering performance. 
In this work, we used SBMs to model \textit{must-link} and \textit{cannot-link} annotations and integrated them into the  minimum sum-of-squares clustering model. We provided efficient learning algorithms and demonstrated that incorporating pairwise information can significantly improve clustering performance, even if the annotations are provided in a small volume and with mistakes. Moreover, for both synthetic and real-world datasets, we have shown that prior knowledge of annotation accuracy can be harnessed to further improve clustering. In challenging cases in which groups overlap substantially and the observed data do not fit the model's general premises, the adoption of pairwise information can be decisive to reveal hidden structures.

This work provides a font of promising research perspectives. Firstly, we suggest exploring other algorithmic approaches for the proposed models along with specific applications such as facial image recognition and video object classification, as discussed in \citet{Basu2008}. Moreover, we suggest other methodological extensions that naturally fit in a semi-supervised framework. Further improvements could also be achieved using active learning to select samples for annotation, or through label propagation techniques~\citep{Zhu2005,Basu2008,Xiong2013}. Finally, this model could be further generalized to consider experts with different accuracy levels and support other probability distributions, increasing its generality and flexibility.



\appendices

\ifCLASSOPTIONcompsoc
  \section*{Acknowledgments}
\else
  \section*{Acknowledgment}
\fi

This research was partially supported by CAPES with Finance Code 001 and grant 88881.189261/2018-01; by CNPq under grants 308528/2018-2 and 140918/2017-5; and by FAPERJ under grant E-26/202.790/2019. This financial support is gratefully acknowledged.

\ifCLASSOPTIONcaptionsoff
  \newpage
\fi



%

%


\bibliographystyle{IEEEtranN}
\bibliography{SSC-IPA}

%




\vspace*{-1.3cm}
\begin{IEEEbiography}[{\includegraphics[width=1in,height=1.25in,clip,keepaspectratio]{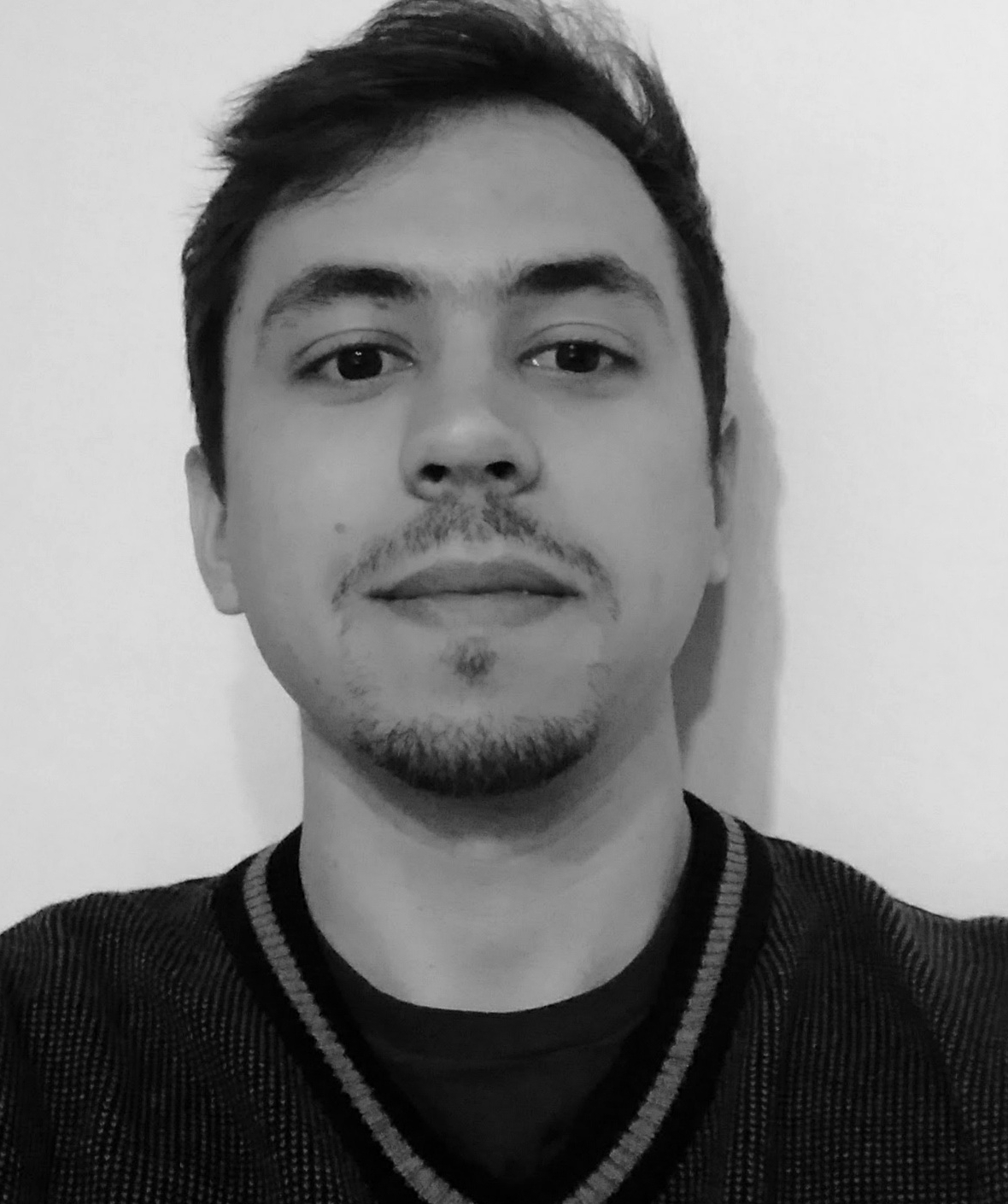}}]{Daniel Gribel}
is currently pursuing a PhD degree in optimization and machine learning in the Department of Computer Science at the Pontifical Catholic University of Rio de Janeiro, Brazil. He is active in a variety of academic and industrial projects. His current research focuses on semi-supervised learning in the presence of noisy information, along with optimization models and methods for clustering and graph partitioning problems.
\end{IEEEbiography}

\vspace*{-1.6cm}
\begin{IEEEbiography}[{\includegraphics[width=1in,height=1.25in,clip,keepaspectratio]{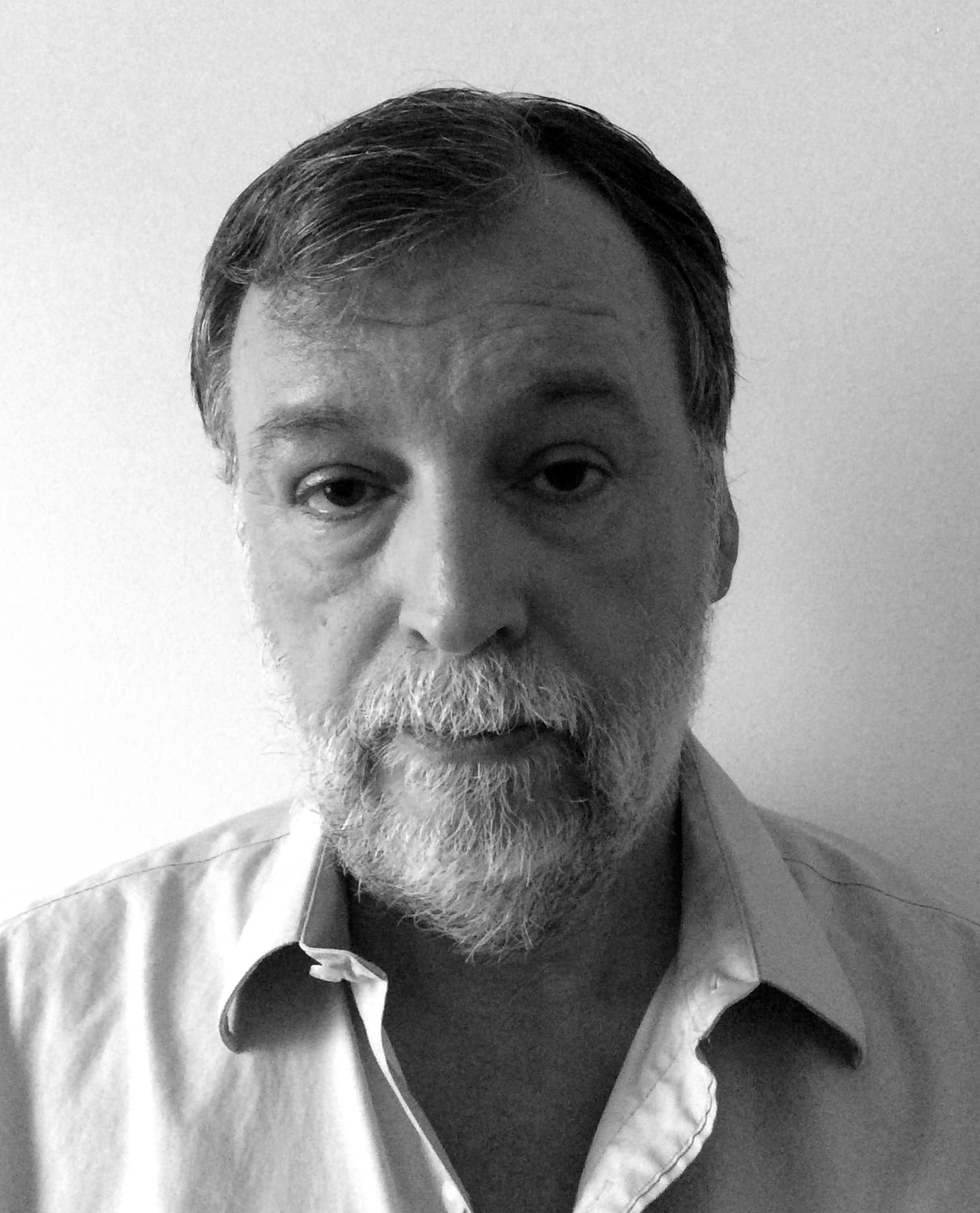}}]{Michel Gendreau} is Professor of Operations Research in the Department of Mathematics and Industrial Engineering of Polytechnique Montréal (Canada). His main research area is the application of operations research methods to transportation and logistics systems planning and operation, energy production and storage, healthcare, and related problems. Dr. Gendreau has published more than 350 papers in peer-reviewed journals and conference proceedings. He is also the co-editor of eight books. Dr. Gendreau was Editor in chief of Transportation Science from 2009 to 2014. In 2015, Dr. Gendreau received the Robert Herman Lifetime Achievement Award of the Transportation Science and Logistics (TSL) Society of INFORMS.
\end{IEEEbiography}

\vspace*{-1.3cm}
\begin{IEEEbiography}[{\includegraphics[width=1in,height=1.25in,clip,keepaspectratio]{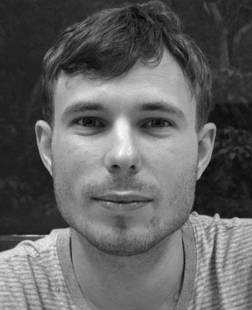}}]{Thibaut Vidal} is a professor at the Department of Mathematics and Industrial Engineering of Polytechnique Montréal, Canada. He is also an adjunct professor at the Department of Computer Science of the Pontifical Catholic University of Rio de Janeiro, Brazil. Dr. Vidal has published over 50 articles in peer-reviewed journals and conference proceedings. He received several awards, including two best paper awards from the TSL Society of INFORMS (2014 and 2016), and the Robert Faure prize from ROADEF (2018). His research interests span over machine learning and combinatorial optimization, with applications to supply chains management, transportation planning, and algorithmic explainability.
\end{IEEEbiography}




\end{document}